\documentclass[letterpaper]{article} 
\usepackage{aaai2026}  
\usepackage{times}  
\usepackage{helvet}  
\usepackage{courier}  
\usepackage[hyphens]{url}  
\usepackage{graphicx} 
\usepackage{natbib}  
\usepackage{caption} 
\usepackage{algorithm}
\usepackage{algorithmic}

\usepackage{newfloat}
\usepackage{listings}
\DeclareCaptionStyle{ruled}{labelfont=normalfont,labelsep=colon,strut=off} 
\floatstyle{ruled}
\newfloat{listing}{tb}{lst}{}
\floatname{listing}{Listing}
\title{\textit{Omni-Effects}: Unified and Spatially-Controllable Visual Effects Generation}
\author {
    Fangyuan Mao\textsuperscript{\rm 1}\equalcontrib,
    Aiming Hao\textsuperscript{\rm 1}\equalcontrib,
    Jintao Chen\textsuperscript{\rm 1,2},
    Dongxia Liu\textsuperscript{\rm 1,3},
    Xiaokun Feng\textsuperscript{\rm 1,4},
    Jiashu Zhu\textsuperscript{\rm 1},
    Meiqi Wu\textsuperscript{\rm 1,4},
    Chubin Chen\textsuperscript{\rm 1,3},
    Jiahong Wu\textsuperscript{\rm 1}\thanks{Corresponding Author},
    Xiangxiang Chu\textsuperscript{\rm 1}
}
\affiliations {
    \textsuperscript{\rm 1}AMAP, Alibaba Group\\
    \textsuperscript{\rm 2}Peking University\\
    \textsuperscript{\rm 3}Tsinghua University\\
    \textsuperscript{\rm 4}Institute of Automation, Chinese Academy of Sciences\\
    fangyuanmaocs@gmail.com, \{haoaiming.ham, hongxi.wjh\}@alibaba.inc.com
}
\usepackage{cuted} 
\usepackage{caption} 
\usepackage{booktabs}
\usepackage{multicol}
\usepackage{multirow}
\usepackage{makecell} 
\usepackage{amsfonts}
\usepackage{pifont}
\usepackage{amsmath}

\begin{document}

\maketitle

\begin{abstract}
Visual effects (VFX) are essential visual enhancements fundamental to modern cinematic production. Although video generation models offer cost-efficient solutions for VFX production, current methods are constrained by per-effect LoRA training, which limits generation to single effects. This fundamental limitation impedes applications that require spatially controllable composite effects, i.e., the concurrent generation of multiple effects at designated locations. However, integrating diverse effects into a unified framework faces major challenges: interference from effect variations and spatial uncontrollability during multi-VFX joint training. To tackle these challenges, we propose \textit{Omni-Effects}, a first unified framework capable of generating prompt-guided effects and spatially controllable composite effects. The core of our framework comprises two key innovations: (1) \textbf{LoRA-based Mixture of Experts (LoRA-MoE)}, which employs a group of expert LoRAs, integrating diverse effects within a unified model while effectively mitigating cross-task interference. (2) \textbf{Spatial-Aware Prompt (SAP)} incorporates spatial mask information into the text token, enabling precise spatial control. Furthermore, we introduce an Independent-Information Flow (IIF) module integrated within the SAP, isolating the control signals corresponding to individual effects to prevent any unwanted blending. To facilitate this research, we construct a comprehensive VFX dataset \textit{Omni-VFX} via a novel data collection pipeline combining image editing and First-Last Frame-to-Video (FLF2V) synthesis, and introduce a dedicated VFX evaluation framework for validating model performance. Extensive experiments demonstrate that \textit{Omni-Effects} achieves precise spatial control and diverse effect generation, enabling users to specify both the category and location of desired effects. 
\end{abstract}
\begin{links}
    \link{Code}{https://amap-ml.github.io/Omni-Effects.github.io}
    \link{Extended version}{https://arxiv.org/abs/2508.07981}
\end{links}

\section{Introduction}

\begin{figure}[t]
    \centering
    \includegraphics[width=1\linewidth]{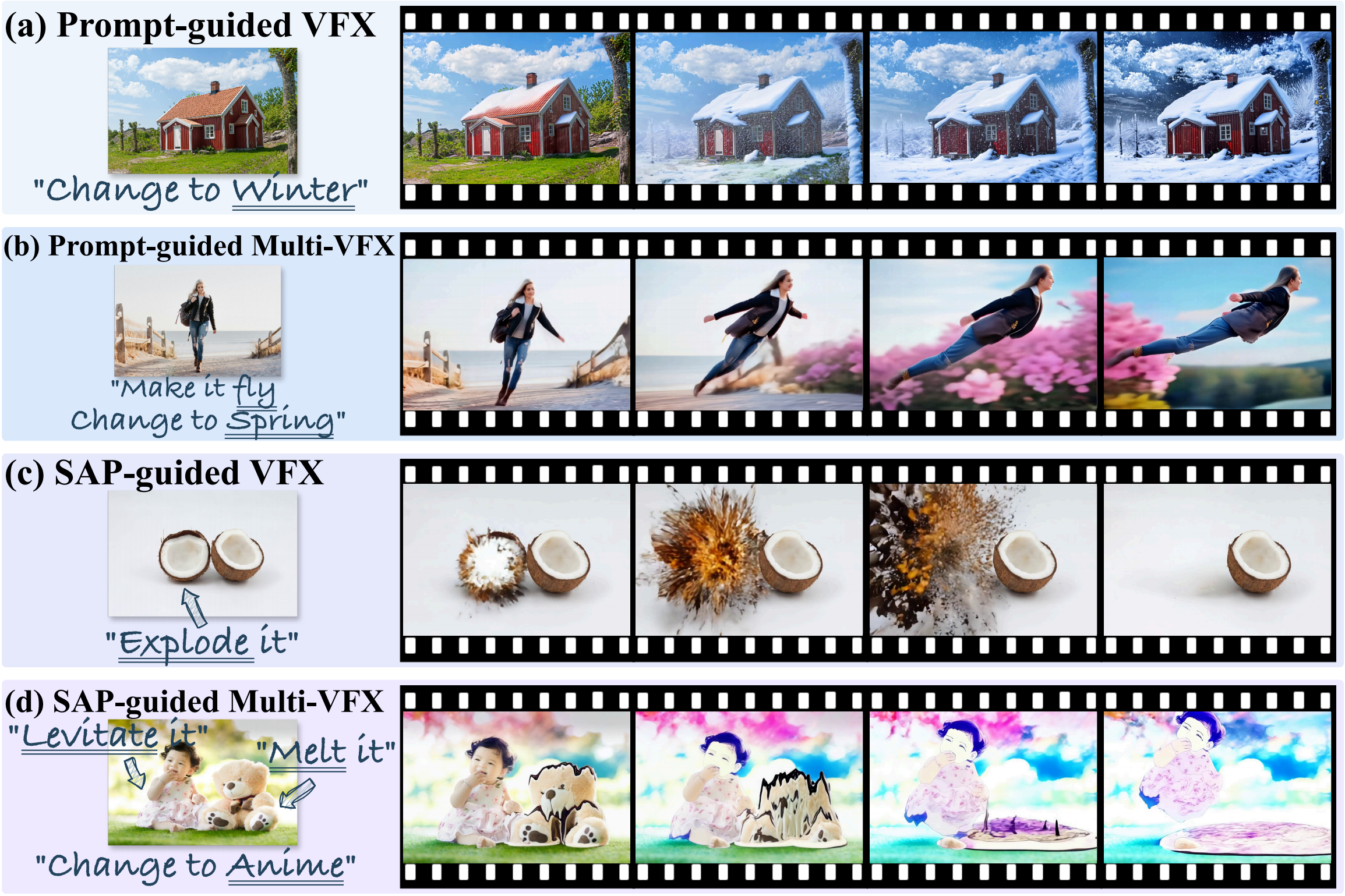}
    \caption{Capabilities for diverse customized visual effects. \textit{Omni-Effects} supports both (a) single-VFX and (b) multi-VFX generation through pure prompt-guided generation. Integrated with the \textbf{Spatial-Aware Prompt}, \textit{Omni-Effects} enables (c) precise spatial VFX control and (d) intricate object-based visual effects with targeted environmental transformations.}
    \label{fig:teaser}
    \vspace{-15pt}
\end{figure}

Visual effects (VFX) play a crucial role in modern filmmaking, enabling the creation of immersive narratives and fantastical worlds. While traditional VFX pipelines, especially for composite effects requiring simultaneous coordination across different spatial locations, are notoriously complex and resource-intensive~\cite{chabanova2022vfx}.  Recent rapid advances in video generation technologies~\cite{Yang2024CogVideoXTD,bao2024vidu,kong2024hunyuanvideo,wan2025} are driving a paradigm shift in VFX creation---transitioning from conventional methods to generative model-powered dynamic, efficient synthesis.

\begin{figure}[t]
    \centering
    \includegraphics[width=1\linewidth]{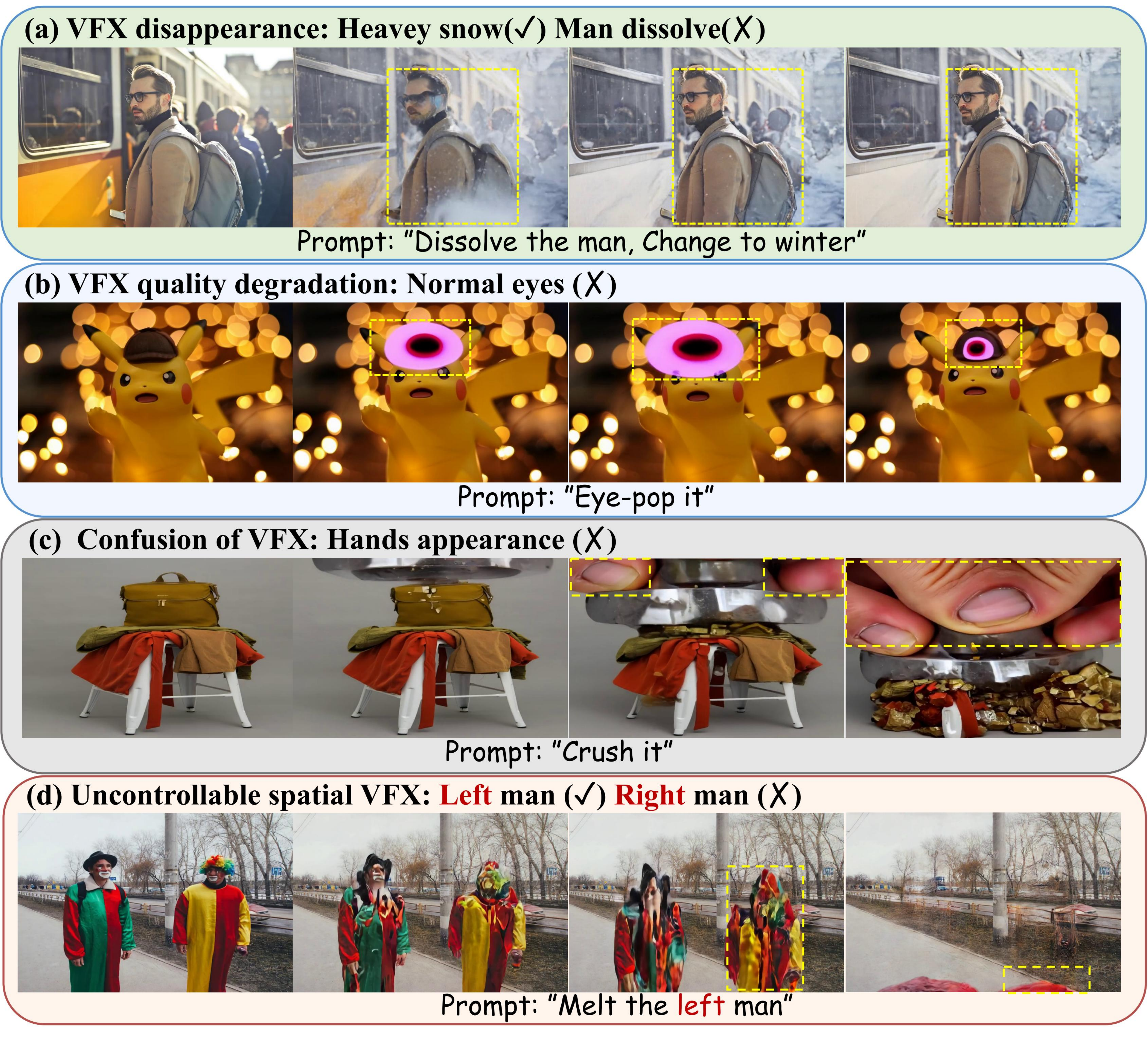}
    \caption{Defects in standard video generation models. (a) VFX disappearance, (b) quality degradation, (c) confusion between VFX elements, and (d) spatial uncontrollability.}
    \label{fig:defects}
\end{figure}

The inherent scarcity of VFX data and pronounced variability in dynamic characteristics across effects pose significant challenges to training generative models. Consequently, current methods~\cite{liu2025vfx} focus on single-effect generation, employing dedicated Low-Rank Adaptation (LoRA)~\cite{hu2022lora} tailored to individual effects. However, this paradigm struggles with multi-VFX scenes, exhibiting two critical limitations: \textbf{Limitation 1}. Cross-Adapter Interference where joint multi-LoRA activation~\cite{ding2023parameter} induces spatial occlusion artifacts (Figure~\ref{fig:defects} (a)) and shared-subspace hybrid training triggers fidelity-degrading cross-effect confusion via task interference~\cite{zhang2025lori} (Figure~\ref{fig:defects} (b, c)). \textbf{Limitation 2}. Spatial-Semantic Misalignment wherein the text-pixel space gap prevents precise spatial cue encoding for VFX placement (Figure~\ref{fig:defects} (d)). These limitations fundamentally constrain conventional video generation adaptation to complex multiple VFX compositions (multi-VFX).

To address these limitations, we propose \textit{Omni-Effects}, a unified framework modeling multi-VFX generation as a multi-condition video generation problem, where textual prompts specify effect categories while spatial masks define their precise locations. First, to tackle Limitation 1, we introduce a \textbf{LoRA-based Mixture of Experts (LoRA-MoE)} module~\cite{shazeer2017outrageously,dou2023loramoe,zhang2025context}, which partitions effects into specialized subspaces, each optimized by dedicated expert branches, with a gating router dynamically activating relevant subspaces to minimize cross-task interference and enhance effect fidelity. Second, to overcome Limitation 2, we present the \textbf{Spatial-Aware Prompt (SAP)}, which integrates explicit mask-based spatial conditioning with textual inputs via full-attention mechanisms for accurate effect placement. To mitigate SAP cross-interference during concurrent application, we introduce the Independent-Information Flow (IIF) module, which isolates condition-specific information flows through IIF Attention Mask, preventing unintended effect blending. Collectively, these innovations enable \textit{Omni-Effects}---to our knowledge, the first VFX framework---to achieve high-fidelity multi-VFX compositions with pixel-level spatial control. (Figure~\ref{fig:teaser}). 

To advance this research, we construct a high-quality VFX dataset \textit{Omni-VFX} and develop a specialized VFX data pipeline. This pipeline utilises image editing models~\cite{liu2025step1x} to generate source image pairs depicting initial/final effect states, which are then synthesised into VFX videos via the FLF2V framework (built on Wan2.1~\cite{wan2025}). Rigorous manual filtering ensures quality while expanding coverage to $55$ distinct effect categories. Further, we introduce an evaluation framework specifically designed for controllable VFX generation tasks. Comprehensive experiments validate the \textit{Omni-Effects} framework's superior performance across three core capabilities: single-VFX, multi-VFX, and controllable VFX generation. In summary, the major contributions of our work are as follows:
\begin{itemize}
    \item A unified VFX framework, \textit{Omni-Effects}, is proposed to enable high-fidelity spatial controllable multi-VFX generation through a dual-core architecture: (1) \textbf{LoRA-MoE} modules unifying multi-VFX training, and (2) IIF-augmented \textbf{SAP} mechanisms enabling independent multi-condition control without interference.
    \item The most comprehensive VFX dataset \textit{Omni-VFX} is developed with an automated production pipeline to support the generation of diverse high-quality VFX video, complemented by a comprehensive evaluation framework for rigorous controllable VFX assessment.
    \item Extensive experiments demonstrate that our \textit{Omni-Effects} achieves precise spatial control and enables diverse VFX generation, thereby allowing users to specify both the category and location of desired effects.
\end{itemize}

\section{Related Works}

\subsection{Video Generation Models}

\begin{figure*}[!t]
    \centering
    \includegraphics[width=1\linewidth]{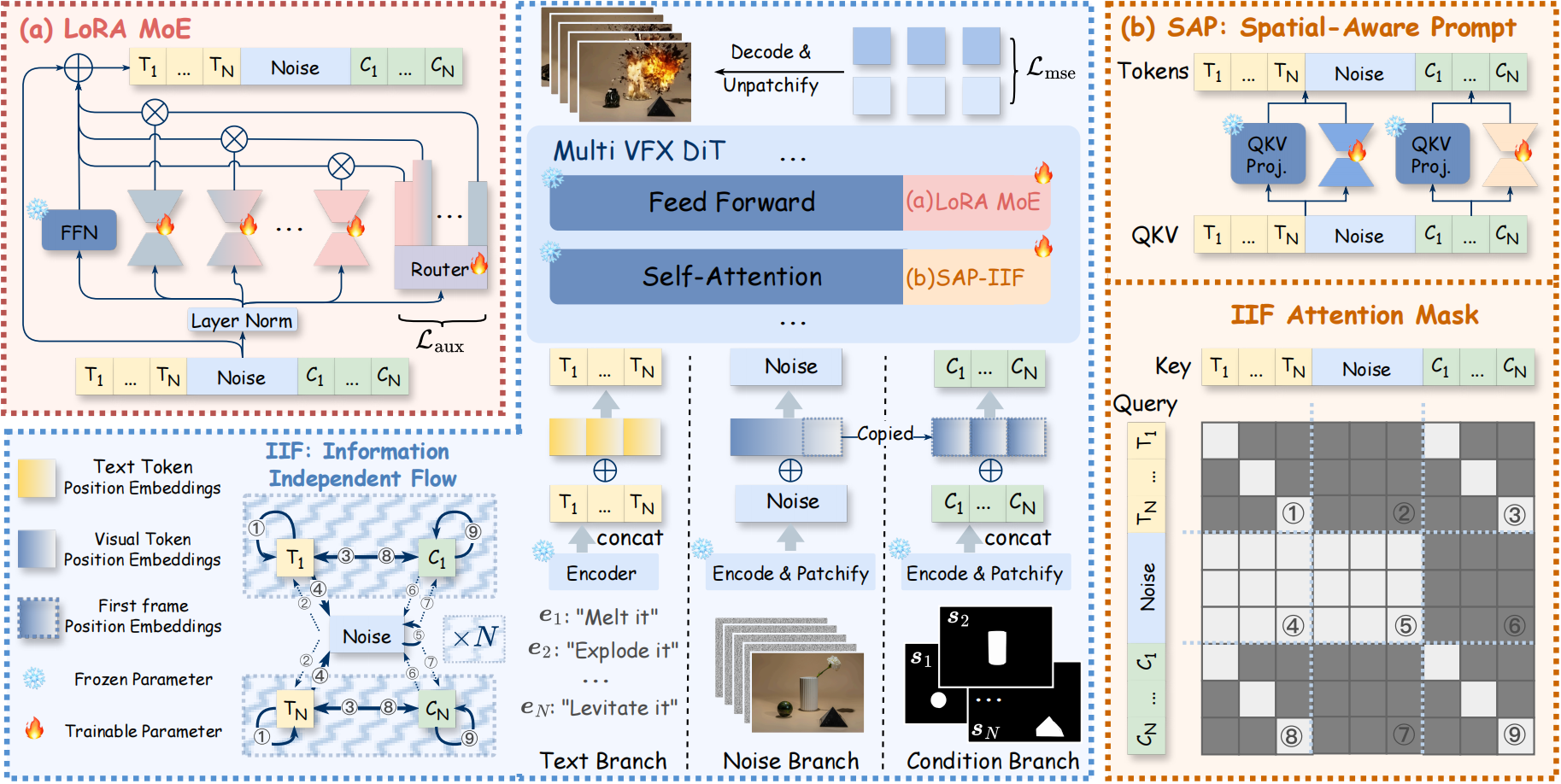}
    \caption{Flowchart of proposed \textit{Omni-Effects}. Given a reference image and composite conditions of arbitrary length, \textit{Omni-Effects} first encodes each input into corresponding tokens. These tokens are concatenated and processed sequentially through downstream DiT blocks. These blocks incorporate two key technologies: (a) \textbf{LoRA-MoE}, a MoE plugin replacing standard FFN linear layers to enable collaborative expert task-solving and (b) \textbf{SAP}, which fuses effect descriptors with spatial trigger information during the attention stage while mitigating cross-condition information leakage via an IIF mechanism. Note that, in the IIF, dashed lines represent blocked information flow, while solid lines indicate active information transmission.}
    \label{fig:framework}
\end{figure*}

Recent advances in diffusion-based video generation~\cite{chen2023videocrafter1,he2022latent,zeng2024make,kong2024hunyuanvideo,Yang2024CogVideoXTD,hacohen2024ltx,bao2024vidu,polyak2024movie,wan2025,teng2025magi,seawead2025seaweed,ling2025vmbench,chen2025s} have enabled text-to-video (T2V) and image-to-video (I2V) synthesis, where input images establish spatial context. As a critical I2V application, VFX generation creates unrealizable fantastical visuals. However, VFX data scarcity forces reliance on Low-Rank Adaptation (LoRA)~\cite{hu2022lora} for limited-data fine-tuning~\cite{liu2025vfx}. This necessitates separate LoRA models per effect, while combined training causes performance degradation, fundamentally limiting multi-VFX generation within a single video.

\textit{Our architecture unifies multi-VFX training, avoids degradation, and enables concurrent multi-VFX generation.}

\subsection{Conditional Video Generation}

Condition-guided diffusion models leverage auxiliary inputs for precise output control, falling into two paradigms: spatial-fusion guidance and high-level semantic guidance. Spatial-fusion methods, exemplified by ControlNet~\cite{Zhang2023AddingCC}, integrate condition inputs with denoising inputs. These methods~\cite{li2024controlnet++,bai2024syncammaster,bian2025videopainter,xu2025hunyuanportrait,lei2025animateanything,jiang2025vace} enable fine-grained spatial alignment while preserving the generation quality of pre-trained diffusion models. High-level semantic methods exploit latent interactions between conditions and the denoising process. Techniques include cross-attention based mechanisms~\cite{ye2023ip, zhang2024ssr,zhou2025light,yuan2025identity} and conditional token concatenation strategies~\cite{wang2024boximator,huang2024context,tan2025ominicontrol,tan2025ominicontrol2,zhang2025easycontrol}, dynamically modulate generation through semantic embeddings. Crucially, while these methods effectively handle individual conditions, they struggle to simultaneously and independently control multiple conditions, a critical requirement for professional VFX generation.

\textit{Our model employs IIF-powered SAP control mechanisms to support independent, non-interfering control of multiple conditions within the same video.} 

\section{Method}

\subsection{Preliminaries}

\subsubsection{Video Diffusion Models}
Video generation models usually utilize the diffusion paradigm~\cite{ho2020denoising,lipman2022flow}, which generates samples from a data distribution $p\left(\boldsymbol{x}_0\right)$ by progressively denoising samples that are initially drawn from a Gaussian distribution $p\left(\boldsymbol{x}_T\right)$. During training, clean samples $\boldsymbol{x}_0 \sim p\left(\boldsymbol{x}_0\right)$ undergo iterative corruption through $T$ diffusion steps:  
\begin{equation}
\begin{aligned}
\boldsymbol{x}_t = \alpha_t \boldsymbol{x}_0 + \sigma_t \boldsymbol{\epsilon}, \quad \boldsymbol{\epsilon} \sim \mathcal{N}\left(\boldsymbol{0},\boldsymbol{I}\right), \ t=1,...,T
\end{aligned}
\end{equation}
where $\alpha_t,\sigma_t>0$ are scalars that jointly define the Signal-to-Noise Ratio (SNR). The denoiser with parameter $\theta$ is optimized to predict the target noise $\boldsymbol{\epsilon}$. The optimization process is defined as:
\begin{equation}
\begin{aligned}
\mathcal{L}_{\text{mse}}\left(\theta\right) = \mathbb{E}_{\boldsymbol{x}_t,t,\boldsymbol{c},\boldsymbol{\epsilon}} \left[\left\| \boldsymbol{\epsilon} - \epsilon_\theta\left(\boldsymbol{x}_t, t,\tau\left(\boldsymbol{c}\right)\right) \right\|_2^2 \right]
\end{aligned}
\end{equation}
where $\boldsymbol{c}$ is conditions (e.g., text and spatial location), and $\tau$ denotes the condition encoder. By replacing target noise $\boldsymbol{\epsilon}$ with $\boldsymbol{v}$, which is a weighted combination of $\boldsymbol{x}_0$ and $\boldsymbol{\epsilon}$, as the prediction target, the v-prediction~\cite{salimans2022progressive} is derived, which is adopted in our \textit{Omni-Effects} framework. Moreover, mainstream video generation leverages Diffusion Transformer (DiT)~\cite{peebles2023scalable} architecture by employing attention mechanisms to model spatiotemporal consistency while aligning conditional inputs with visual outputs~\cite{zheng2024open}. By integrating diffusion processes with Transformer architectures, the video generation performance is improved, leading to high-quality and accurate video synthesis results.

\subsubsection{Spatially Controllable Multi-VFX Generation}
In practical applications, it's often necessary to display different VFX at distinct locations throughout a video. We formalize this task as multi-conditional video generation, wherein video diffusion models take a reference image and a set of $N$ control signals $\boldsymbol{C}=\left\{\boldsymbol{c}_i\right\}_{i=1}^N$ as inputs. Each condition $\boldsymbol{c}_i=\left(\boldsymbol{e}_i,\boldsymbol{s}_i\right)$ couples an effect descriptor $\boldsymbol{e}_i$ with a spatial trigger $\boldsymbol{s}_i$, whereby the generated video $\boldsymbol{x}_0$ applies effect $\boldsymbol{e}_i$ at the location specified by $\boldsymbol{s}_i$. Specifically, we use a text prompt to describe the VFX, while using a spatial mask $\boldsymbol{m}\in \mathbb{R}^{H\times W}$ to serve as the spatial trigger. A set of conditions is incorporated into the denoising process, and the denoiser prediction becomes:
\begin{equation}
\begin{aligned}
\hat{\boldsymbol{v}}= \epsilon_\theta\left(\boldsymbol{x}_t, t,\left\{\tau_{e}^{\left(i\right)}\left(\boldsymbol{e}_i\right)\right\}_{i=1}^N,\left\{\tau_{s}^{\left(i\right)}\left(\boldsymbol{s}_i\right)\right\}_{i=1}^N\right),
\end{aligned}
\end{equation}
where $\tau_{e}$ and $\tau_{s}$ denote the text and spatial mask encoder, respectively. Notice that, when $N=1$ and the spatial trigger is empty, the above task reduces to the traditional single-VFX generation task~\cite{liu2025vfx}.

\subsection{\textit{Omni-Effects}}
To model the above task, video diffusion models require simultaneous support for multi-VFX inference and spatial control capabilities. We accordingly propose \textit{Omni-Effects}, building upon the CogVideoX~\cite{Yang2024CogVideoXTD} architecture and incorporating two core components: \textbf{LoRA-MoE} and \textbf{Spatial-Aware Prompt}. The overview is illustrated in Figure~\ref{fig:framework}, and the details are as follows.

\begin{figure}[!t]
    \centering
    \includegraphics[width=1\linewidth]{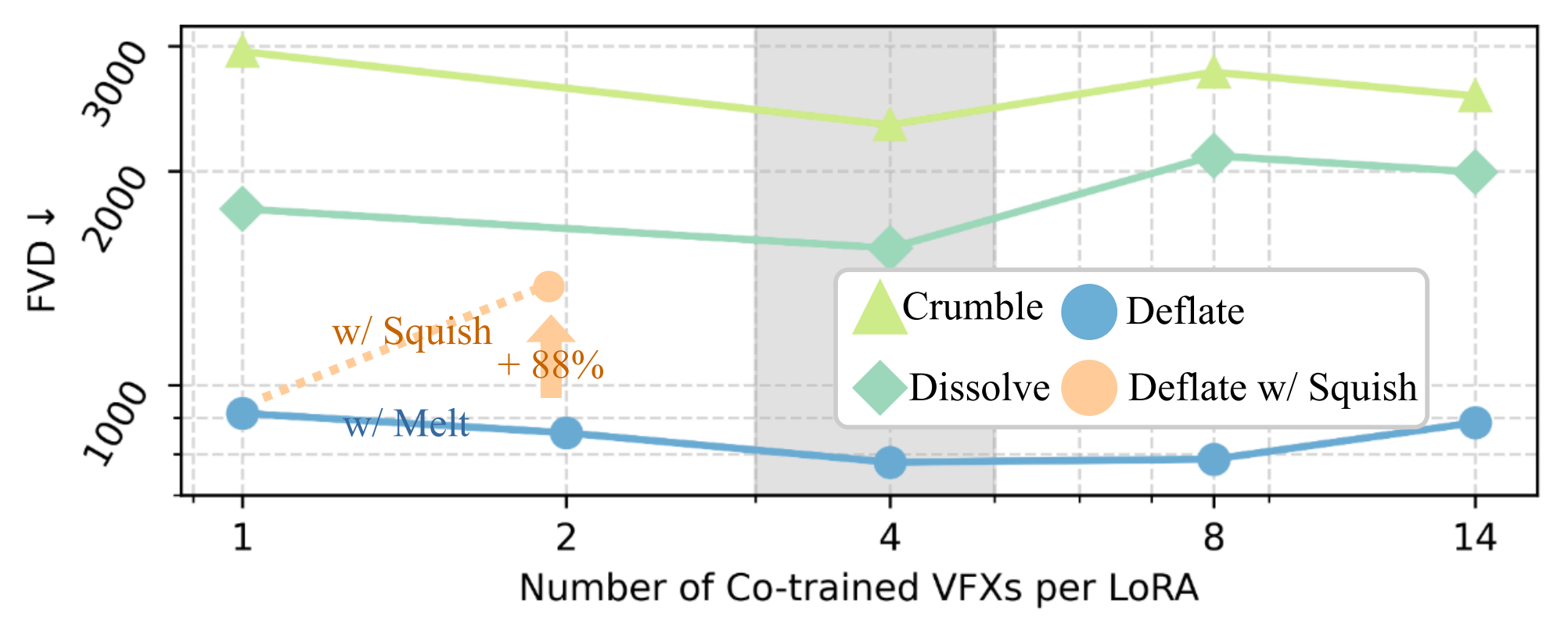}
    \caption{FVD scores for diverse VFX trained with a shared LoRA. VFX performance exhibits an initial improvement followed by progressive degradation with increasing numbers of co-trained effects. This indicates inherent effect clustering: synergistic groups (e.g., Melt-like effects) improve co-training performance, while incompatible combinations (e.g., Deflate + Squish) suffer from mode collapse and underperform relative to compatible sets. Note that, \textbf{lower FVD values indicate superior performance}, with optimal VFX results uniformly achieved when the number of co-trained VFX equals 4.}
    \label{fig:fvd}
\end{figure}

\subsubsection{LoRA-MoE}

As mentioned in Figure~\ref{fig:defects}, both multi-LoRA parallel inference and single-LoRA unified training degrade performance. Crucially, we observe the synergistic mechanism in VFX training: \textit{compatible VFX-combination training enhances single-VFX generation quality} (Figure~\ref{fig:fvd}). This discovery motivates our adaptive task-space partitioning strategy: inspired by MoE~\cite{shazeer2017outrageously} architectures, we partition distinct effects into specialized subspaces and deploy a gating router for adaptive subspace selection.

Specifically, LoRA-MoE~\cite{dou2023loramoe} integrates MoE with LoRA (Figure~\ref{fig:framework} (a)), which employs an expert ensemble where each LoRA specializes in distinct VFX manifolds. Formally, for input token $\boldsymbol{x}\in \mathbb{R}^d$, a weight is obtained by a gating network $G:\mathbb{R}^d\mapsto \mathbb{R}^n$ for each expert, resulting in $G\left(\boldsymbol{x}\right)=\left[G\left(\boldsymbol{x}\right)_1,G\left(\boldsymbol{x}\right)_2,...,G\left(\boldsymbol{x}\right)_n\right]$, where $n$ represents the number of experts. Each expert $E_i$ implements LoRA decomposition: 
\begin{equation}
E_i(\boldsymbol{x}) = \frac{\alpha}{r}\boldsymbol{x} \boldsymbol{A}_i \boldsymbol{B}_i,\ \boldsymbol{A}_i \in \mathbb{R}^{d \times r},\ \ \boldsymbol{B}_i \in \mathbb{R}^{r \times d},
\end{equation}
where $r$ denotes the low-rank and $\alpha$ is a scaling factor. The final prediction combines base model and expert outputs:
\begin{equation}
\begin{aligned}
\boldsymbol{y} =\text{Base}\left(\boldsymbol{x}\right)+ \sum_{i=1}^nG\left(\boldsymbol{x}\right)_i\odot E_i\left(\boldsymbol{x}\right),
\end{aligned}
\end{equation}
This MoE-structured plugin replaces standard FFN linear layers, enabling collaborative expert task-solving. During training, Top-$k$ routing strategy ($k\le n$) is adopted to enforce exactly $k$ non-zero entries in $G\left(\boldsymbol{x}\right)$. At inference, all experts are activated to prevent effect suppression caused by Top-$k$ filtering, omitting critical experts, which is essential for multi-VFX combination generation. Moreover, to mitigate workload imbalance caused by the gating network favoring a few experts during training, we also employ a balanced routing auxiliary loss $\mathcal{L}_{\text{aux}}$~\cite{fedus2022switch}. Comprehensive details are provided in Supplement \textbf{A}. The final training objective is expressed as: $\mathcal{L}=\mathcal{L}_{\text{mse}}+\beta\mathcal{L}_{\text{aux}}$, where $\beta$ is hyperparameter.


\begin{figure}[!t]
    \centering
     \includegraphics[width=1\linewidth]{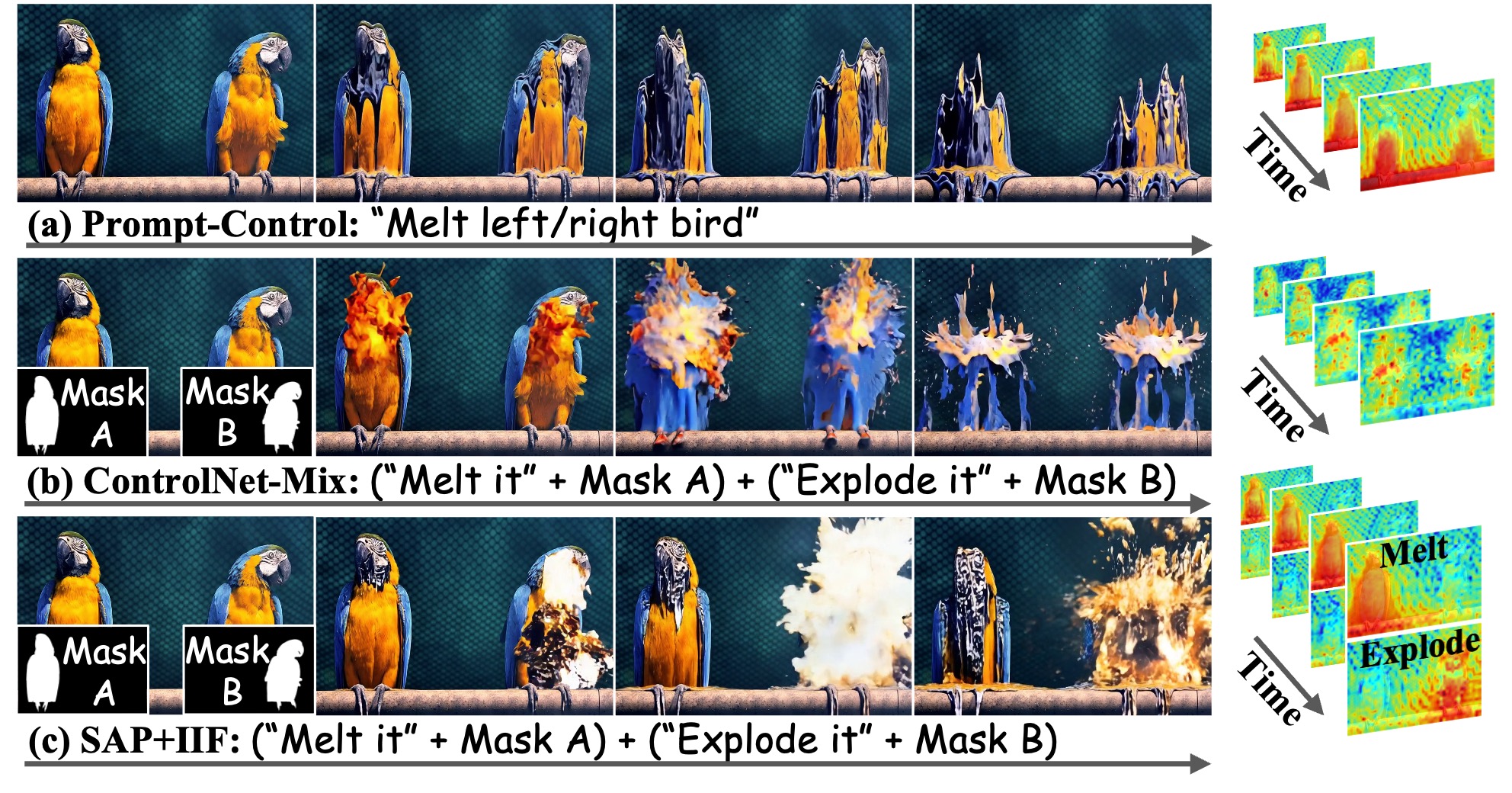}
    \caption{Visualization of controllable VFX performance and attention maps. (a) Position description lacks spatial control; (b) ControlNet faces inter-condition interference, leading to VFX leakage and artifacts; (c) Proposed \textbf{SAP+IIF} achieves precise positional controllability while preventing mutual interference between multi-VFX.}
    \label{fig:control}
\end{figure}

\subsubsection{Spatial-Aware Prompt} 
For condition $\boldsymbol{c}_i=\left(\boldsymbol{e}_i,\boldsymbol{s}_i\right)$, embedding positional descriptors within text prompts proves insufficient for precise spatial control. To investigate this phenomenon, we visualize attention maps across diverse prompts. Crucially, \textit{attention consistently activates identical regions regardless of prompt semantics} (Figure~\ref{fig:control} (a)), evidencing textual position cues' failure to direct activation toward specified targets. Prior work~\cite{liu2025vfx,jiang2025vace} mitigates this via ControlNet~\cite{Zhang2023AddingCC} to extract a mask sequence for generation guidance. However, this solution suffers from two critical limitations:
\begin{enumerate}
    \item \textbf{Significant parameter overhead}: ControlNet duplicates a portion of the base model's parameters (typically half), requiring substantial extra trainable weights;
    \item \textbf{Severe cross-condition interference}: During multi-VFX generation, parallel ControlNet inference suffers from information leakage, manifesting as erroneous co-occurrence of effect $\boldsymbol{e}_i$ and $\boldsymbol{e}_j$ at positions $\boldsymbol{s}_i$ and $\boldsymbol{s}_j$ respectively (Figure~\ref{fig:control} (b)).
\end{enumerate}

In summary, signals within composite conditions must be integrated while preventing cross-condition interference to ensure robust performance. We address these challenges by proposing the \textbf{Spatial-Aware Prompt} to directly inject spatial information into prompts tokens via enhanced spatial-text condition token interactions within attention mechanism, enabling controllable generation with minimal parameter/computational overhead. Building on this, we introduce Information-Independent Flow, which utilizes a designed attention mask to restrict cross-condition information exchange, thereby preventing interference between distinct control streams. Formally, given a set of conditions $\boldsymbol{C}$, encoder processing yields text condition tokens $\left\{\tau_{e}^{\left(i\right)}\left(\boldsymbol{e}_i\right)\right\}_{i=1}^N$ and spatial condition tokens $\left\{\tau_{s}^{\left(i\right)}\left(\boldsymbol{s}_i\right)\right\}_{i=1}^N$, which are sequentially concatenated with the noisy latent $\boldsymbol{x}_t$ to form the inputs $\boldsymbol{Q}$, $\boldsymbol{K}$ and $\boldsymbol{V}$. Then we define an attention mask $\boldsymbol{M}\in \left\{0,-\infty\right\}^{l\times l}$ ($l$ is the total sequence length) to regulate attention flow (Figure~\ref{fig:framework} (b), details are in Supplement \textbf{A}) that blocks condition-to-condition and noise-to-condition interactions, eliminating cross-condition leakage to prevent effect misalignment or blending. The final output of attention is expressed as:
\begin{equation}    \boldsymbol{y}=\text{Softmax}\left(\boldsymbol{Q}\boldsymbol{K}^T/\sqrt{d_k}+\boldsymbol{M}\right)\boldsymbol{V},
\end{equation}
where $d_k$ denotes the feature dimension. To enhance spatial conditioning alignment with noisy latents, we inject positional embeddings from $\boldsymbol{x}_t$'s initial frame into $\tau_{s}^{\left(i\right)}\left(\boldsymbol{s}_i\right)$, coupled with a dedicated Spatial-Condition LoRA. Crucially, all spatial conditions share identical LoRA parameters while maintaining a common base LoRA across other branches, ensuring efficient conditional injection without disrupting pretrained representations. Each text condition is individually processed through text encoder while sharing identical positional encoding. As shown in Figure~\ref{fig:control} (c), our SAP+IIF achieves precise VFX targeting in target regions with non-overlapping activation zones.

\section{Data and Training}

\subsection{Dataset Collection}

VFX fundamentally manifest as radical spatio-temporal state transformations (e.g., explosion). Despite modern techniques like animation and Computer Graphics Interface (CGI)~\cite{chabanova2022vfx}, modeling such dynamics remains challenging. We introduce a novel pipeline: for any input image, Step1X-Edit~\cite{liu2025step1x} produces its modified counterpart to establish boundary frames defining a VFX's initial and terminal states. This constraint provides strong transformation priors for generative models. The FLF2V framework~\cite{wan2025} then synthesizes the final video by compressing VFX production into boundary-constrained state-transition path search, significantly reducing modeling complexity. Through curated manual selection, we build a comprehensive dataset \textit{Omni-VFX} spanning \textbf{55 distinct VFX} across \textbf{instantaneous environmental shifts, artistic styles, human emotions, and so on}, enabling diverse creative applications. For more data details, please refer to Supplement \textbf{B}.

\subsection{Training}

Since our training dataset contains only single-VFX without multi-VFX data, empirical observations reveal that standard training fails to achieve controllable multi-VFX generation. We overcome this with a tri-level solution. At the \textbf{data} level, through random cropping and splicing with two videos, and random temporal freezing, we generate pseudo multi-VFX videos with corresponding masks. At the \textbf{scheduler} level, Non-Uniform Sampling prioritizes denoising steps $\in \left[900,1000\right]$(early stage) for spatial control learning with increased batch allocation, while dedicating fewer batches to detail refinement in lower steps $\in \left[0,900\right]$, motivated by empirical findings that enhanced focus on early denoising accelerates model convergence. At the \textbf{training strategy} level, iterative single to multi-VFX ($N=2$) fine-tuning ensures stable convergence and performance gains. For more training details, please refer to Supplement \textbf{C}.

\begin{table*}[!ht]
\scriptsize
\setlength\tabcolsep{2pt}
\begin{tabular}{@{}cccccccccccccccccc@{}}
\toprule
\textbf{Metrics} & \textbf{Methods} & \textbf{Cake-ify} & \textbf{Crumble} & \textbf{Crush} & \textbf{Decapitate} & \textbf{Deflate} & \textbf{Dissolve} & \textbf{Eye-pop} & \textbf{Harley} & \textbf{Inflate} & \textbf{Levitate} & \textbf{Melt} & \textbf{Squish} & \textbf{Ta-da} & \textbf{Venom} & \textbf{Avg.} & \textbf{Param.$^\#$} \\ \midrule
\multirow{3}{*}{\textbf{FVD}$\downarrow$} 
& \textbf{Single LoRA} & 2138 & 2947 & 1496 & \underline{1190} & 913 & \underline{1770} & 1995 & \underline{3576} & \underline{1505} & 1401 & 2827 & \underline{1415} & \textbf{1053} & 4146 & \underline{2026} & 132.1M \\
& \textbf{Mix LoRA} & \underline{1674} & 2552 & 1772 & 1299 & \underline{886} & 1995 & \underline{1725} & 4496 & 2042 & \underline{1006} & \underline{2748} & \textbf{1225} & 1240 & \underline{3923} & 2041 & 9.4M \\
 \cmidrule(lr){2-18} 
& \textbf{LoRA-MoE} & \textbf{1506} & \textbf{1641} & \textbf{1213} & \textbf{1177} & \textbf{839} & \textbf{1118} & \textbf{1460} & \textbf{3330} & \textbf{1304} & \textbf{736} & \textbf{2512} & 1561 & \underline{1064} & \textbf{3339} & \textbf{1628} & 28.5M \\ \midrule
\multirow{3}{*}{\makecell[c]{\textbf{Dynamic}\\\textbf{Degree}}$\uparrow$} & \textbf{Single LoRA} & 0.8 & 0.8 & 0.0 & 0.6 & 0.0 & \textbf{0.8} & 0.0 & 1.0 & 0.8 & 0.0 & 0.6 & 1.0 & 1.0 & 1.0 & 0.60 & 132.1M \\
 & \textbf{Mix LoRA} & 0.8 & 0.8 & 0.0 & 0.6 & 0.0 & \textbf{0.8} & 0.0 & 1.0 & 0.8 & 0.0 & 0.6 & 1.0 & 1.0 & 1.0 & 0.60 & 9.4M \\
 \cmidrule(lr){2-18} 
 & \textbf{LoRA-MoE} & \textbf{1.0} & \textbf{1.0} & \textbf{0.6} & \textbf{0.6} & 0.0 & 0.4 & 0.0 & \textbf{1.0} & \textbf{1.0} & 0.0 & \textbf{0.6} & \textbf{1.0} & \textbf{1.0} & \textbf{1.0} & \textbf{0.66} & 28.5M \\ 
 \bottomrule
\end{tabular}
\caption{Performance comparison on OpenVFX dataset. \textbf{Param.$^\#$} represents the average training parameters per effect. And the highest metric values are highlighted in \textbf{bold}, with the second-best \underline{underlined}.}
\label{tab:lora-moe}
\end{table*}

\begin{table*}[!ht]
\centering
\setlength{\tabcolsep}{1.5mm}
\begin{tabular}{lcccccccccccc}
\toprule
\multirow{2}{*}{\textbf{Methods}} & \multicolumn{4}{c}{\textbf{RDD}$\uparrow$} & \multicolumn{4}{c}{\textbf{EOR}$\uparrow$} & \multicolumn{4}{c}{\textbf{ECR}$\uparrow$} \\ 
\cmidrule(lr){2-5} \cmidrule(lr){6-9}  \cmidrule(lr){10-13} 
 & \textbf{Melt} & \textbf{Levitate} & \textbf{Explode} & \textbf{Avg.} & \textbf{Melt} & \textbf{Levitate} & \textbf{Explode} & \textbf{Avg.} & \textbf{Melt} & \textbf{Levitate} & \textbf{Explode} & \textbf{Avg.} \\ \midrule
\textbf{CogVideoX} & 0.91 & 0.99 & 1.11 & 1.00 & 0.06 & 0.09 & 0.11 & 0.09 & 0.00 & 0.00 & 0.00 & 0.00 \\
\textbf{LTX-Video} & 0.12 & 0.11 & 0.14 & 0.12 & 0.05 & 0.02 & 0.05 & 0.04 & 0.00 & 0.00 & 0.00 & 0.00 \\
\textbf{Wan-2.1} & 2.06 & 1.57 & 2.38 & 2.00 & 0.11 & 0.02 & 0.03 & 0.05 & 0.02 & 0.00 & 0.00 & 0.01 \\
\textbf{CogV+CN} & \textbf{3.80} & \textbf{2.39} & \underline{2.09} & \underline{2.76} & \underline{0.95} & \underline{0.80} & \underline{0.82} & \underline{0.86} & \underline{0.56} & \underline{0.36} & \underline{0.70} & \underline{0.54} \\ \midrule
\textbf{Ours} & \underline{2.69} & \underline{2.22} & \textbf{3.87} & \textbf{2.93} & \textbf{0.99} & \textbf{0.94} & \textbf{0.99} & \textbf{0.97} & \textbf{0.93} & \textbf{0.83} & \textbf{0.89} & \textbf{0.88} \\ \bottomrule
\end{tabular}
\caption{Quantitative results of Single-VFX control generation. We compare \textit{Omni-Effects} with representative open-source video generation models under three controllable VFX scenarios: Melt, Levitate, and Explode.}
\label{tab:single-control}
\end{table*}

\section{Experiments}

\subsection{Experimental Setup}

\subsubsection{Evaluation Metrics}
Following previous work~\cite{liu2025vfx}, for single-VFX evaluation, we employ two established metrics: Fréchet Video Distance (FVD)~\cite{unterthiner2018towards} for overall fidelity and Dynamic Degree~\cite{huang2024vbench} for motion dynamics. For controllable VFX, we introduce three novel metrics. \emph{Regional Dynamic Degree} (RDD), which utilizes optical flow and masks, quantifies the strength of motion within the target region, thereby quantifies motion strength within target regions to measure visual impression. \emph{Effect Occurrence Rate} (EOR), which is computed by inputting both the video and a given prompt template into Gemini2.5~\cite{comanici2025gemini} to obtain the answer, measures intended effect trigger frequency, indicating generation reliability. Building upon EOR, \emph{Effect Controllability Rate} (ECR) assesses spatial precision by verifying VFX confinement to designated areas. Complete metric details appear in Supplement \textbf{D}.

\subsubsection{Implementation Details}

Training employs a CogVideoX-5B backbone with LoRA rank of $128$ with a total of $n=8$ experts, generating $49\times480\times720$ resolution videos. For loss, $\beta$ is set to $0.01$. We utilize 8 H20 GPUs (96GB) with a batch size of 1 per GPU. We use AdamW~\cite{loshchilov2017decoupled} at a constant $10^{-4}$ learning rate for $5,000$ steps. During inference, DDIM~\cite{nichol2021improved} sampling ($50$ steps) with CFG~\cite{ho2022classifier} scale $6.0$ is applied, which can be perform on a single GPU. Extended details are in Supplement~\textbf{C}.

\subsection{Quantitative Results}

In the following, we evaluate the effectiveness of \textit{Omni-Effects} by comparing it with baseline models on unified and controllable VFX generation tasks.

\subsubsection{Unified VFX Generation} 
We evaluate the LoRA-MoE against VFX-specific training LoRA and mix training LoRA for all VFX on the public OpenVFX dataset as detailed in Table~\ref{tab:lora-moe}. LoRA-MoE achieves the best performance across different types of VFX, while significantly reducing the number of trainable parameters. This demonstrates the effectiveness of the designed VFX task-subspace partitioning strategy. Qualitative results are shown in Supplement \textbf{E}.

\subsubsection{Controllable VFX Generation}

To evaluate our model, we perform comprehensive experiments for single- and multi-VFX control, comparing with state-of-the-art methods including (a) CogVideoX~\cite{Yang2024CogVideoXTD}, (b) LTX-Video~\cite{hacohen2024ltx}, (c) Wan2.1~\cite{wan2025}, and (d) CogVideoX integrated with ControlNet (CogV+CN). Evaluation targets three spatially localized VFX types---Explode, Melt, and Levitate---to ensure contamination-free assessment.

\textbf{Single-VFX Control.} Table~\ref{tab:single-control} demonstrates baseline methods' fundamental limitations in synthesizing target VFX and achieving precise spatial control. While CogV+CN can synthesize VFX, it exhibits limited controllability. In comparison, \textit{Omni-Effects} achieves the best performance with \textbf{0.97} EOR and \textbf{0.88} ECR, significantly outperforming all baselines in both generation quality and spatial control precision. This validates that our proposed SAP effectively integrates VFX descriptors with spatial triggers without introducing substantial additional training parameters. Qualitative comparisons are shown in Supplement \textbf{E}.

\begin{figure}[!t]
    \centering
    \includegraphics[width=1\linewidth]{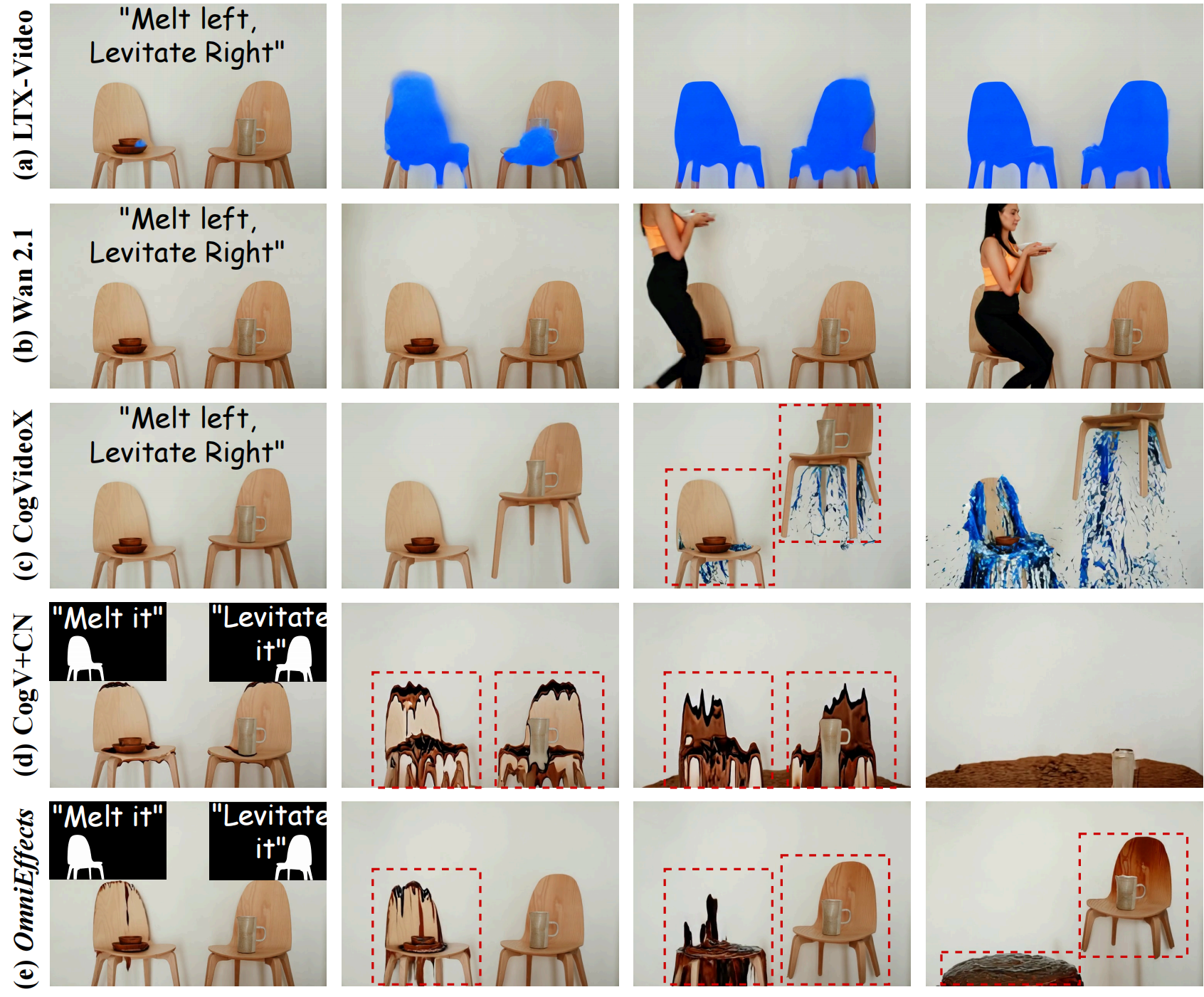}
    \caption{Qualitative comparison of Multi-VFX generation. The desired outcome requires the left chair to melt while the right levitates simultaneously.}
    \label{fig:multi-vfx-qualitative}
\end{figure}

\textbf{Multi-VFX Control.} For multi-VFX generation using two effects combinations, Table~\ref{tab:multi-control} shows baseline models consistently failing to generate or spatially control VFX. \textit{Omni-Effects} achieves precise spatial control over simultaneous VFX. Moreover, Figure~\ref{fig:multi-vfx-qualitative} demonstrates \textit{Omni-Effects}' superiority: when instructed to melt the left chair while levitating the right chair, CogVideoX erroneously applies melting to both objects; CogV+CN correctly renders melting but fails to generate levitation; whereas \textit{Omni-Effects} simultaneously executes both VFX through spatial condition. This performance validates our proposed IIF's efficacy in mitigating cross-condition interference. The user study is shown in Supplement \textbf{E}.

\begin{table}[!t]
\centering
\setlength{\tabcolsep}{0.5mm}
\begin{tabular}{lcccccc}
\toprule
\multirow{2}{*}{\textbf{Methods}} & \multicolumn{3}{c}{\textbf{Melt+Levitate}} & \multicolumn{3}{c}{\textbf{Melt+Explode}} \\ 
 \cmidrule(lr){2-4} \cmidrule(lr){5-7}
 \cmidrule(lr){2-4} \cmidrule(lr){5-7}
 & \begin{tabular}[c]{@{}c@{}}\textbf{RDD}$\uparrow$\end{tabular} & \begin{tabular}[c]{@{}c@{}}\textbf{EOR}$\uparrow$\end{tabular} & \begin{tabular}[c]{@{}c@{}}\textbf{ECR}$\uparrow$\end{tabular} & \begin{tabular}[c]{@{}c@{}}\textbf{RDD}$\uparrow$\end{tabular} & \begin{tabular}[c]{@{}c@{}}\textbf{EOR}$\uparrow$\end{tabular} & \begin{tabular}[c]{@{}c@{}}\textbf{ECR}$\uparrow$\end{tabular} \\ \midrule 
\textbf{CogVideoX}  & 1.80 & 0.00 & 0.00 & 0.96 & 0.03 & 0.00 \\
\textbf{LTX-Video} & 0.11 & 0.00 & 0.00 & 0.13 & 0.00 & 0.00 \\
\textbf{Wan-2.1}   & 1.93 & 0.01 & 0.00 & 3.10 & 0.01 & 0.00 \\
\textbf{CogV+CN} & \textbf{3.18} & 0.09 & 0.05 & \underline{3.60} & 0.08 & 0.08 \\ \midrule
\textbf{Ours} & \underline{2.63} & \textbf{0.68} & \textbf{0.41} & \textbf{4.59} & \textbf{0.62} & \textbf{0.50} \\ \bottomrule
\end{tabular}
\caption{Quantitative results of Multi-VFX generation. \textit{Omni-Effects} achieves a high success rate of independent control over Multi-VFX.}
\label{tab:multi-control}
\end{table}

\begin{figure}[!t]
    \centering
    \includegraphics[width=1\linewidth]{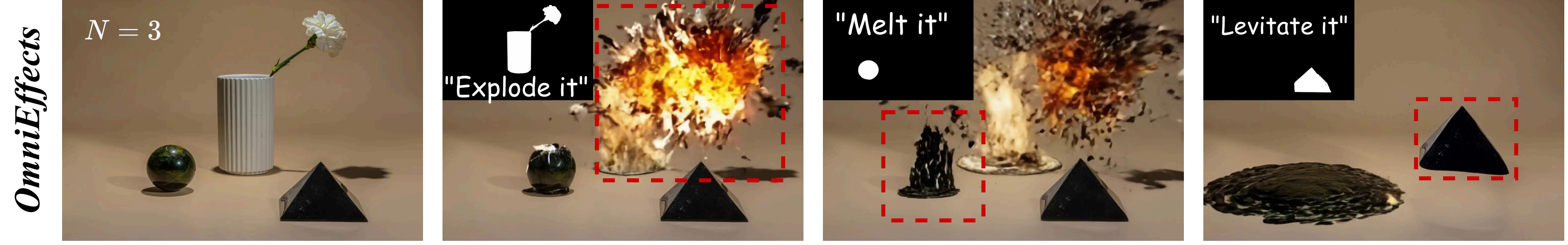}
    \caption{Scalable VFX augmentation. \textit{Omni-Effects} supports inference-time extension to diverse VFX composition.}
    \label{fig:3obj}
\end{figure}

\subsubsection{Generalization} 

Despite being trained with only $N=2$ effects, our model generalizes to diverse mask conditions during inference using the shared Spatial-Condition LoRA, thereby extending to the generation of more concurrent control VFX ($N>2$). \textit{Omni-Effects} demonstrates robust extensibility, successfully handling complex effect combinations (Figures~\ref{fig:teaser} (d) and ~\ref{fig:3obj}), validating its test-time scalable VFX control capability.

\subsection{Ablation Studies}

\subsubsection{LoRA-MoE}

Ablation study on expert count $n$ and Top-$k$ selection (Table~\ref{tab:ablation-moe}) reveals that scaling experts improves generation quality at increased parameter cost. Crucially, our MoE architecture with minimal experts surpasses LoRA baselines (Table~\ref{tab:lora-moe}), demonstrating efficient VFX adaptation through parameter-optimized expert aggregation.

\subsubsection{SAP+IIF}

\begin{figure}[!t]
    \centering
    \includegraphics[width=1\linewidth]{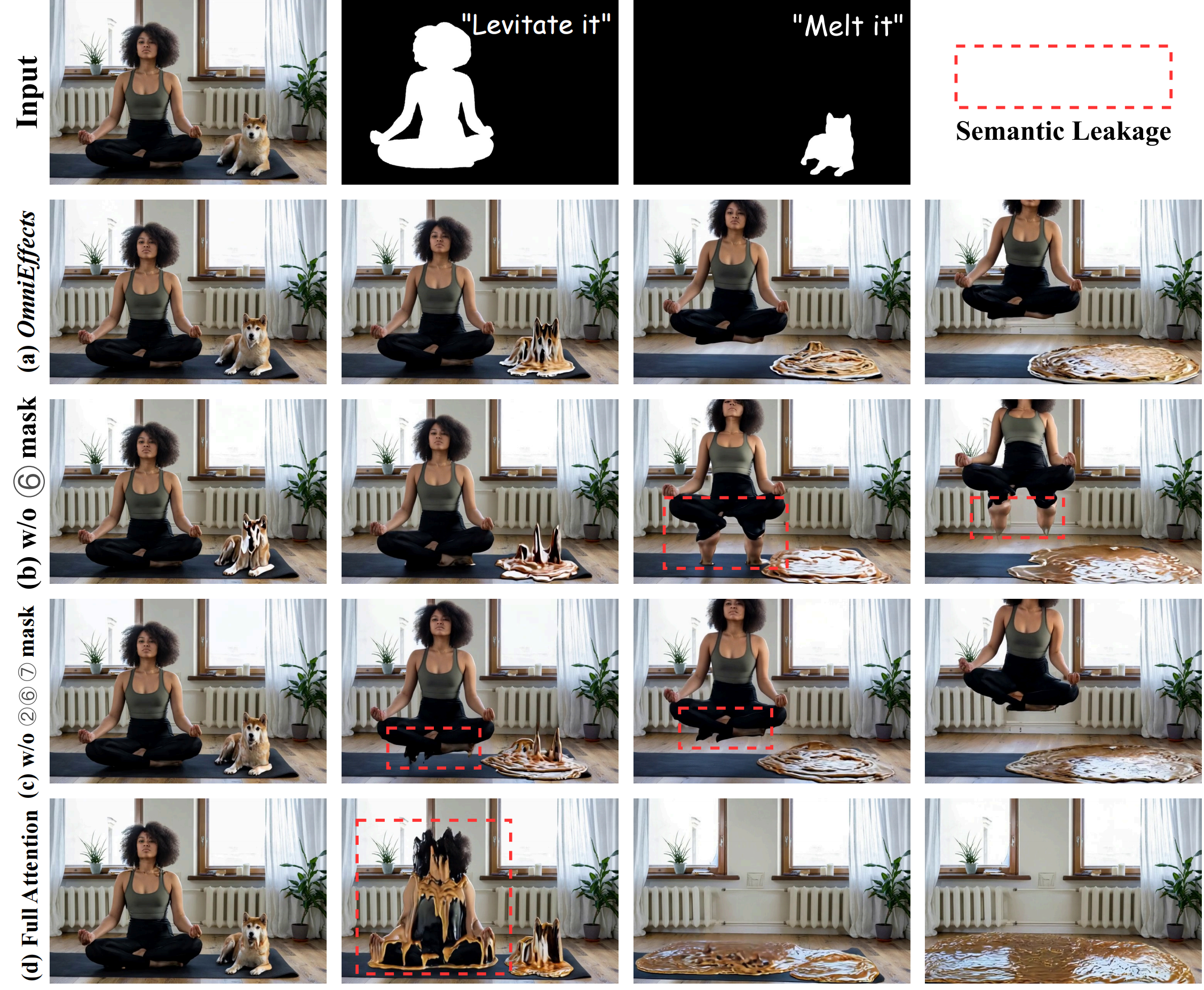}
    \caption{Effect of different attention masks in SAP. Attention Masks are progressively removed while information flow constraints are relaxed from top to bottom.}
    \label{fig:ablation-attn}
\end{figure}

Ablation study on SAP+IIF reveal critical insights in attention mechanisms corresponding to information flow. Removing SAP attention masks from regions \{\ding{193},\ding{197},\ding{198}\} causes melting artifacts on levitating objects (Figure~\ref{fig:ablation-attn} (b, c)), exposing information leakage, while complete attention induces uncontrolled object melting, demonstrating excessive information interaction degrades control. Strategic masking of \{\ding{193},\ding{197},\ding{198}\} prevents leakage while preserving independent information flow in target regions. Additional ablation studies are detailed in the Supplement \textbf{F}.

\begin{table}[!t]
\centering
\begin{tabular}{@{}cccc@{}}
\toprule
\textbf{Metrics} & \textbf{Model} & \textbf{Avg.} & \textbf{Param.$^\#$} \\ \midrule
\multirow{2}{*}{\textbf{FVD}$\downarrow$} & 4 Experts+Top1 & 1762 & 18.9 \\
 & 8 Experts+Top2 & \textbf{1628} & 28.5 \\ \midrule
\multirow{2}{*}{\begin{tabular}[c]{@{}c@{}}\textbf{Dynamic}\\ \textbf{Degree}\end{tabular}$\uparrow$} & 4 Experts+Top1 & 0.65 & 18.9 \\
 & 8 Experts+Top2 & \textbf{0.66} & 28.5 \\ \bottomrule 
\end{tabular}
\caption{Ablation study on LoRA MoE settings. Scaling experts improves generation quality, at the expense of more parameters.}
\label{tab:ablation-moe}
\end{table}

\section{Conclusion}

In this paper, we propose \textit{Omni-Effects}, a unified framework for generating customized VFX videos. It supports the creation of diverse VFX, ranging from single-VFX, multi-VFX to spatially controllable multi-VFX. To achieve these, our framework integrates two core modules: LoRA-MoE and SAP-IIF. Specifically, the LoRA-MoE module mitigates cross-condition interference arising during mix training of multi-VFX. The SAP module, on the other hand, fuses VFX descriptors with spatial trigger information and tackles cross-condition information leakage via an IIF mechanism. Through the synergistic integration of LoRA-MoE and SAP-IIF, \textit{Omni-Effects} enables precise spatial control and produces high-fidelity multi-VFX composites. We also develop a comprehensive VFX dataset \textit{Omni-VFX} with a specialized data production pipeline and an evaluation framework tailored for controllable VFX generation to further validate our approach. Extensive experiments demonstrate the robustness of \textit{Omni-Effects} across complex, multi-condition VFX generation scenarios. Multi-VFX generation represents a domain of substantial practical value coupled with persistent technical challenges. To the best of our knowledge, this work pioneers the first comprehensive framework explicitly addressing this complex problem. Our methodology substantively advances controllable multi-VFX synthesis capabilities while unlocking novel applications across film production, game development, and advertising creatives.

\bibliography{aaai2026}

\clearpage
\appendix
\setcounter{page}{1}

\maketitle

\section{Method}
\subsection{Balanced Routing Auxiliary Loss $\mathcal{L}_\text{aux}$}
Drawing inspiration from Switch Transformers~\cite{fedus2022switch}, we integrate a balanced routing auxiliary loss $\mathcal{L}_{\text{aux}}$ into LoRA MoE training. Specifically, for a batch $\mathcal{B}$ with $\mathcal{T}$ tokens, define $f_i$ as the fraction of tokens routed to expert $i$:
\begin{equation}
\begin{aligned}
f_i=\frac{1}{\mathcal{T}}\sum_{\boldsymbol{x}\in \mathcal{B}}\mathbb{I}\left\{\arg\max p\left(\boldsymbol{x}\right)=i\right\},
\end{aligned}
\end{equation}
and $P_i$ is the mean router probability for expert $i$: 
\begin{equation}
\begin{aligned}
P_i=\frac{1}{\mathcal{T}}\sum_{\boldsymbol{x}\in \mathcal{B}}{p_i\left(\boldsymbol{x}\right)}.
\end{aligned}
\end{equation}
The auxiliary loss encourages load balancing through:
\begin{equation}
\begin{aligned}
\mathcal{L}_{\text{aux}}=n\sum_{i=1}^nf_i\cdot P_i,
\end{aligned}
\end{equation}
which reaches its theoretical minimum of $1$ when $p_i(\boldsymbol{x}) = 1/n$ uniformly ($\forall \boldsymbol{x}, i$).
When the gating network outputs an average probability distribution of $[1/n,\cdots,1/n]$ for tokens in a batch, $\mathcal{L}_\text{aux}$ achieves its minimum value as $n\sum_{i=1}^n{1/n\cdot1/n}=1$

\subsection{IIF Attention Mask}


The IIF Attention Mask is partitioned into two primary components: 
\begin{itemize}
\item \textbf{Condition Interaction}: Within each condition pair $\boldsymbol{c}_i$, comprising text ($\tau_{e}^{\left(i\right)}\left(\boldsymbol{e}_i\right)$) and spatial condition ($\tau_{s}^{\left(i\right)}\left(\boldsymbol{s}_i\right)$) tokens, tokens attend to each other. However, tokens within one condition pair are masked from all tokens in other condition pairs and from the noisy latent token 
$\boldsymbol{x}_t$ to prevent information leakage. 
\item \textbf{Information Aggregation}: The noisy latent token $\boldsymbol{x}_t$ attends to all text tokens ($\left\{\tau_{e}^{\left(k\right)}\left(\boldsymbol{e}_k\right)\right\}_{k=1}^N$) and to itself. This aggregates textual information for updating its representation. Crucially, $\boldsymbol{x}_t$ is masked from all spatial condition tokens ($\left\{\tau_{s}^{\left(k\right)}\left(\boldsymbol{s}_k\right)\right\}_{k=1}^N$), preventing direct access and avoiding redundancy. 
\end{itemize}
Formally, the IIF Attention Mask $M_{ij}$ for tokens $x_i$ and $x_j$ is defined as:
\begin{equation}
M_{ij} = \begin{cases}
0, & \text{if } x_i \text{ and } x_j \text{ belong to the same condition } \boldsymbol{c}_k \\
0, & \text{if } x_i \in \boldsymbol{x}_t \text{ and } x_j \in \boldsymbol{x}_t \cup \left\{\tau_{e}^{\left(k\right)}\left(\boldsymbol{e}_k\right)\right\}_{k=1}^N \\
-\infty, & \text{otherwise}
\end{cases}
\end{equation}

\section{Dataset}
\subsection{Dataset Collection}

To augment our dataset, we employ three strategies: a novel \textbf{first-last frame generation} method and integrating external datasets. 

\begin{figure*}
    \centering
     \includegraphics[width=1\linewidth]{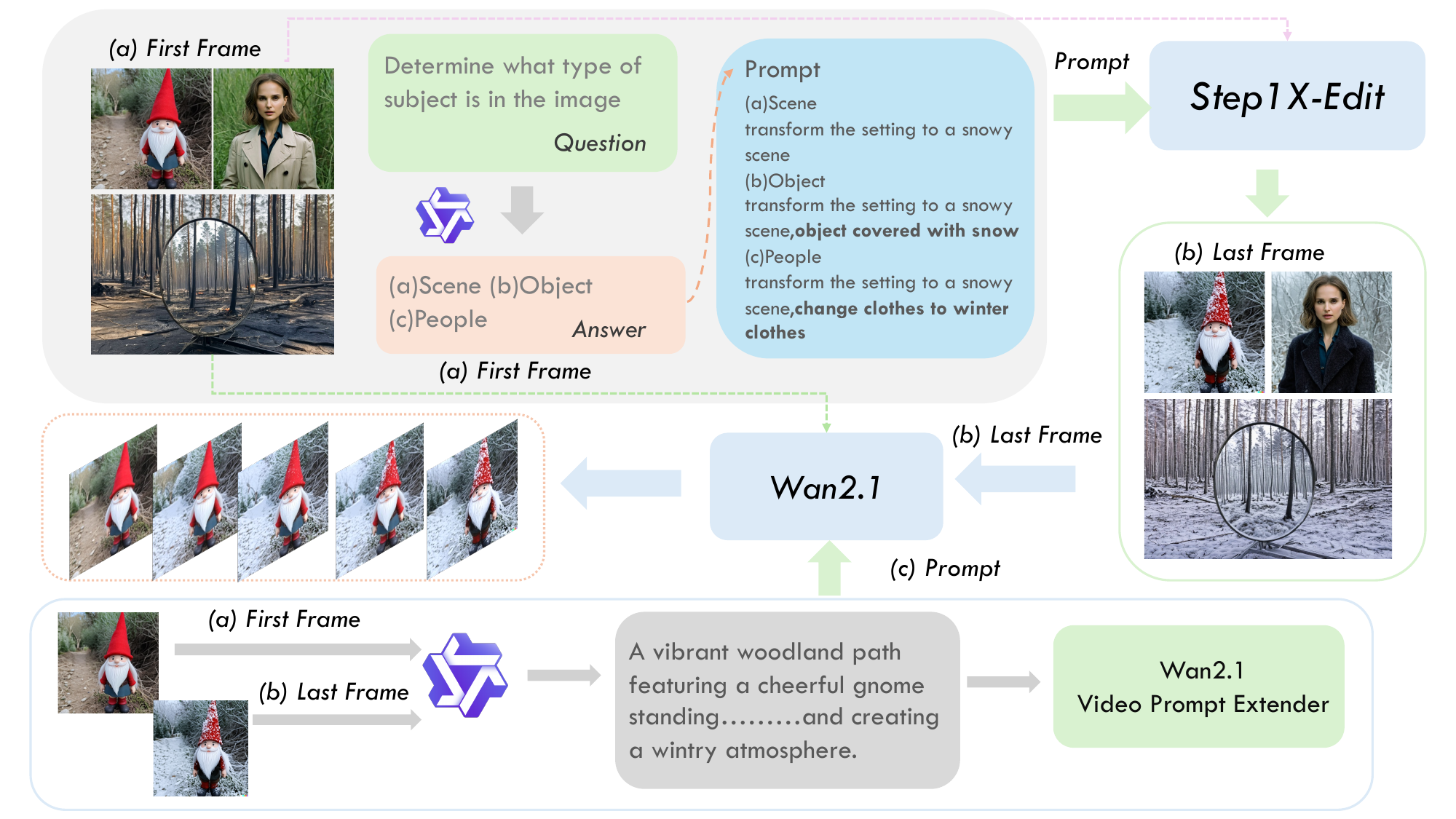 }
    \caption{\textbf{Synthetic VFX Video Generation via Keyframe Editing and WAN 2.1 Interpolation.}}
    \label{fig:dataset_overview}
\end{figure*}

Our dataset construction pipeline, outlined in Figure~\ref{fig:dataset_overview}, follows a multi-stage generative approach. \textbf{Firstly}, we define target VFX categories spanning diverse styles (e.g., seasonal transformations, claymation, 3D doll rendering). An initial semantic analysis using Qwen~\cite{qwen} classifies input images by content and style. \textbf{Secondly}, for each category, Step1X-Edit~\cite{liu2025step1x} generates stylistically consistent frame pairs (initial-final) conditioned on dynamically constructed prompts. These pairs are then analyzed by Qwen to produce descriptive text prompts, which undergo iterative refinement through Wan2.1-14B's Video Prompt Extender~\cite{wan2025} for temporal coherence. \textbf{Thirdly}, the optimized prompts and frames jointly drive Wan2.1-14B to synthesize augmented video sequences. Through this pipeline, we achieve coverage of 55 effect categories while maintaining quality via: (1) automated style-consistency checks, and (2) manual validation of visual fidelity.

\begin{figure*}
    \centering
     \includegraphics[width=1\linewidth]{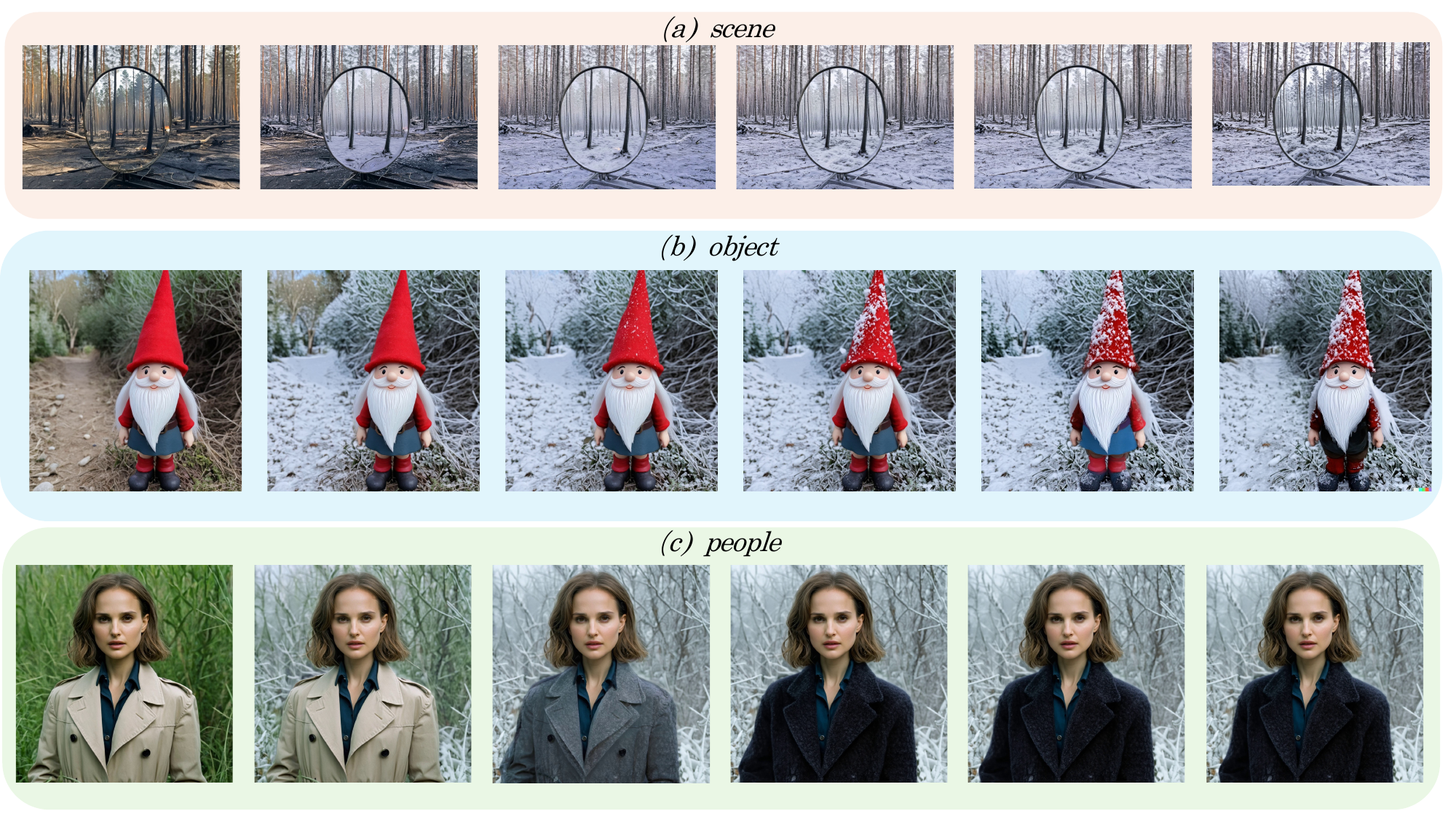 }
    \caption{\textbf{Some examples of our dataset curation pipeline.}}
    \label{fig:dataset_result}
\end{figure*}

\begin{figure*}
    \centering
     \includegraphics[width=1\linewidth]{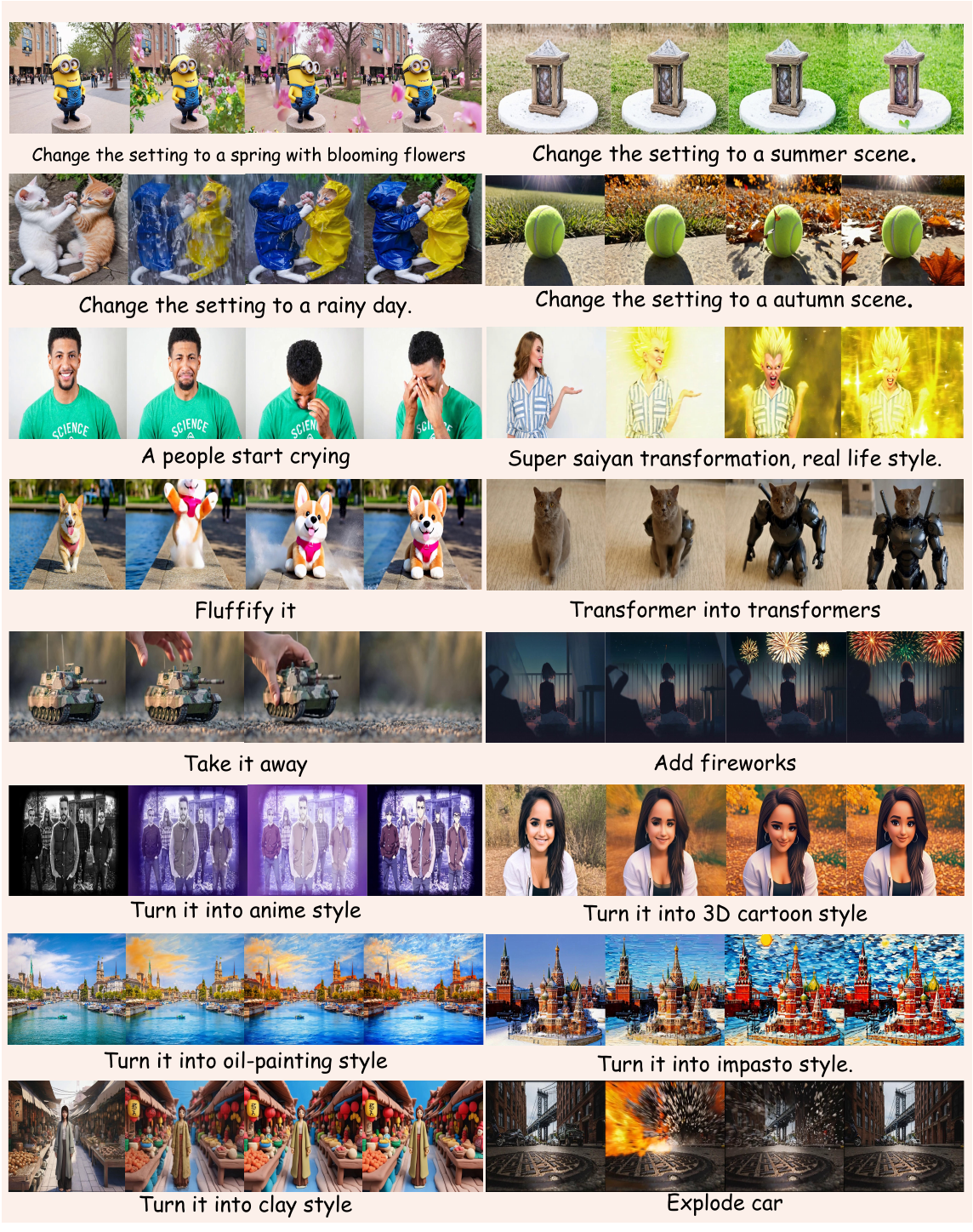}
    \caption{\textbf{Some examples of our \textit{Omni-VFX} dataset.}}
    \label{fig:visualization}
\end{figure*}

\subsection{Dataset Visualization}

Our dataset consists of 55 diverse VFX samples, systematically categorized into five groups: 
\begin{enumerate}
    \item \textbf{Environmental Shifts}: Including seasonal transitions (spring, summer, autumn, winter) and weather variations (e.g., rain).  
    \item \textbf{Dynamic Transformations}: Featuring simulated phenomena such as explosions and fluid dynamics.  
    \item \textbf{Artistic Styles}: Comprising stylized renderings (e.g., oil painting effects, 3D doll aesthetics).  
    \item \textbf{Human Emotion Depictions}: Capturing facial expressions (e.g., smiles, crying).  
    \item \textbf{Complex Effects}: Integrating multiple visual elements across categories to produce sophisticated composites. 
\end{enumerate}

\begin{figure}
    \centering
     \includegraphics[width=1\linewidth]{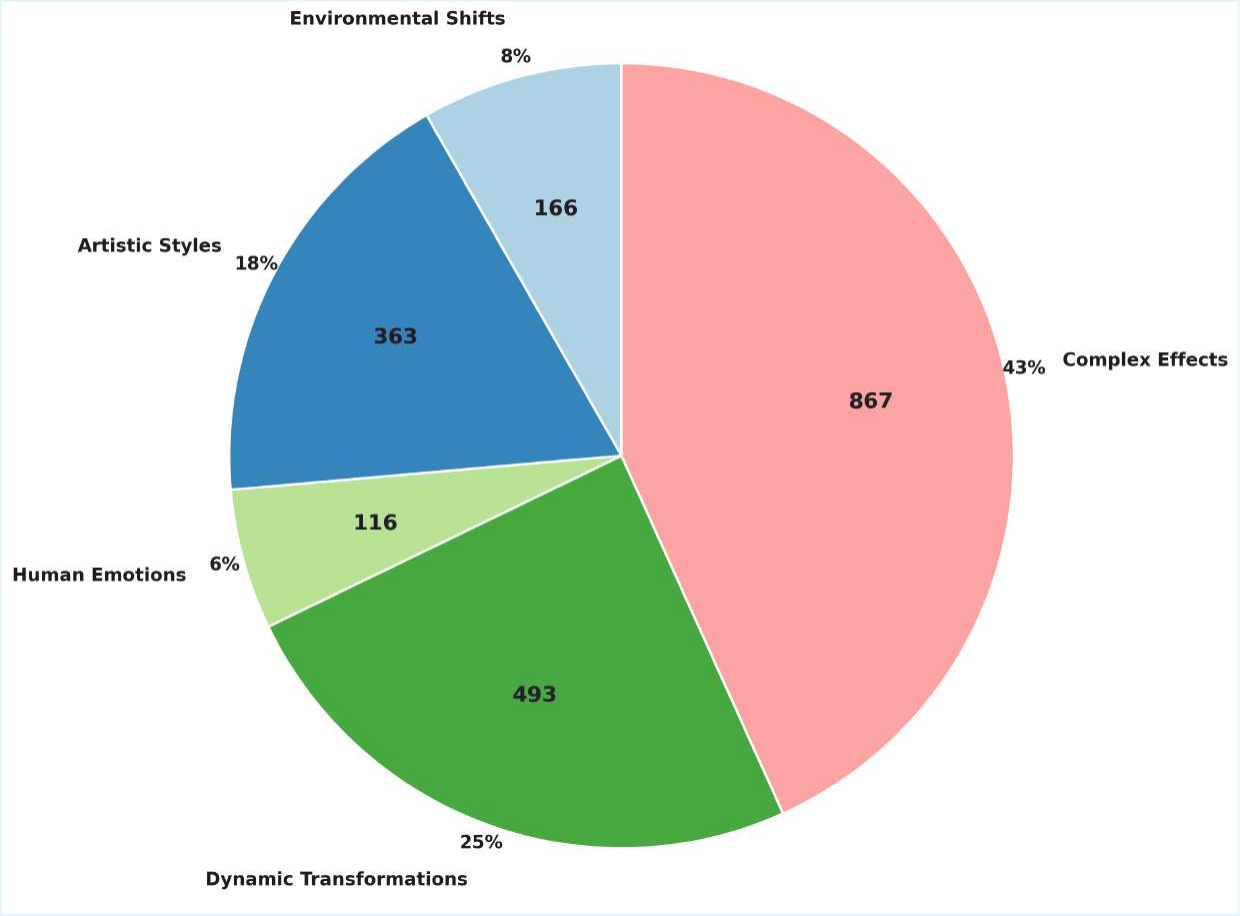}
    \caption{\textbf{Distribution of our \textit{Omni-VFX} dataset.}}
    \label{fig:distribute}
\end{figure}

The distribution of samples across categories is illustrated in Figure~\ref{fig:distribute}, while representative visualizations are provided in Figure~\ref{fig:visualization}.

\section{Implementation Details}

\subsection{Training Data Augmentation}

\begin{figure}
    \centering
    \includegraphics[width=1\linewidth]{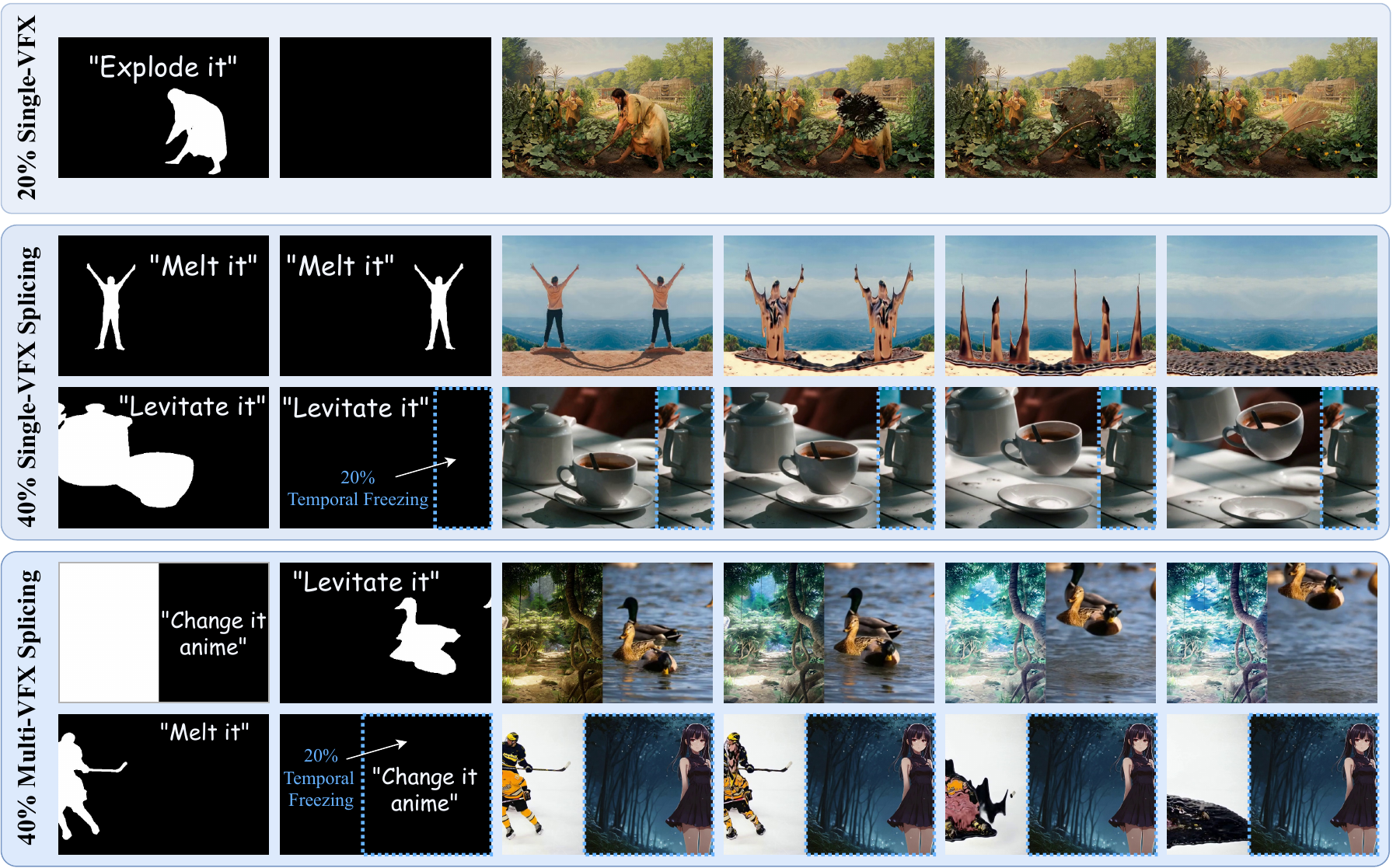}
    \caption{\textbf{Visualization of Data augmentation.}}
    \label{fig:data_aug}
\end{figure}

To address the scarcity of multi-VFX (Multiple Visual Effects) data, we perform data augmentation on single-VFX data. We sample single-VFX data along with their corresponding complete mask data with a probability of 20\%; with a 40\% probability, we sample single-VFX data and perform random cropping and splicing; and with a 40\% probability, we sample two types of VFX and perform random cropping and splicing. During the splicing process, there is a 20\% probability of applying temporal freezing to any segment of the spliced video (turning it into a static video with the corresponding mask set to empty, simulating the condition where no VFX is generated). The example of data augmentation is shown in Figure~\ref{fig:data_aug}.

\subsection{Non-Uniform Timestep Sampling}

Through experiments, we observe that for video generation with strict control requirements, the denoising steps in the early stages of the diffusion model play a critical role, as they determine whether the target region achieves precise control. Traditional uniform timestep sampling requires extensive training to sufficiently optimize the model's accuracy in the initial steps. To accelerate training convergence, we enhance the focus on early denoising by allocating $3/4$ of the batch to the crucial initial steps ($[900, 1000]$), while dedicating the remaining $1/4$ to refining details in later steps ($[0, 900]$). 

\subsection{Dual-phase Training Strategy}

To enhance the model's capability in controlling both single-VFX and multi-VFX videos' generation, we adopt a dual-stage progressive training strategy during the training phase:
\begin{itemize}
    \item \textbf{Stage 1}: We train the model using single-VFX videos with single masks, allowing it to learn how to control a single VFX. This stage runs for $2,000$ steps.  
    \item \textbf{Stage 2}: In addition to single-VFX videos, we introduce multi-VFX videos by combining two VFXs (as described in Sec. C.1) and perform data augmentation on these samples. The model is fine-tuned for an additional 3,000 steps under this setting.
\end{itemize}

This training approach improves the model's robustness and enables it to generalize to a larger number of control VFXs ($N > 2$) during inference.  

\begin{figure*}[!t]
    \centering
     \includegraphics[width=1\linewidth]{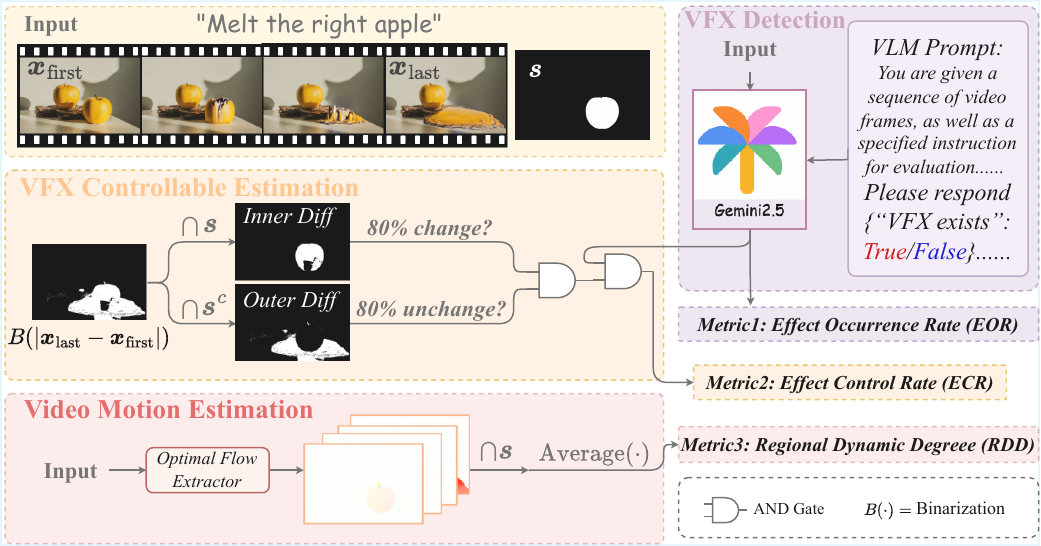}
    \caption{\textbf{Flowchart of the metric design for controllable visual effect.}}
    \label{fig:metrics}
\end{figure*}

\begin{figure*}[!t]
    \centering
    \includegraphics[width=1\linewidth]{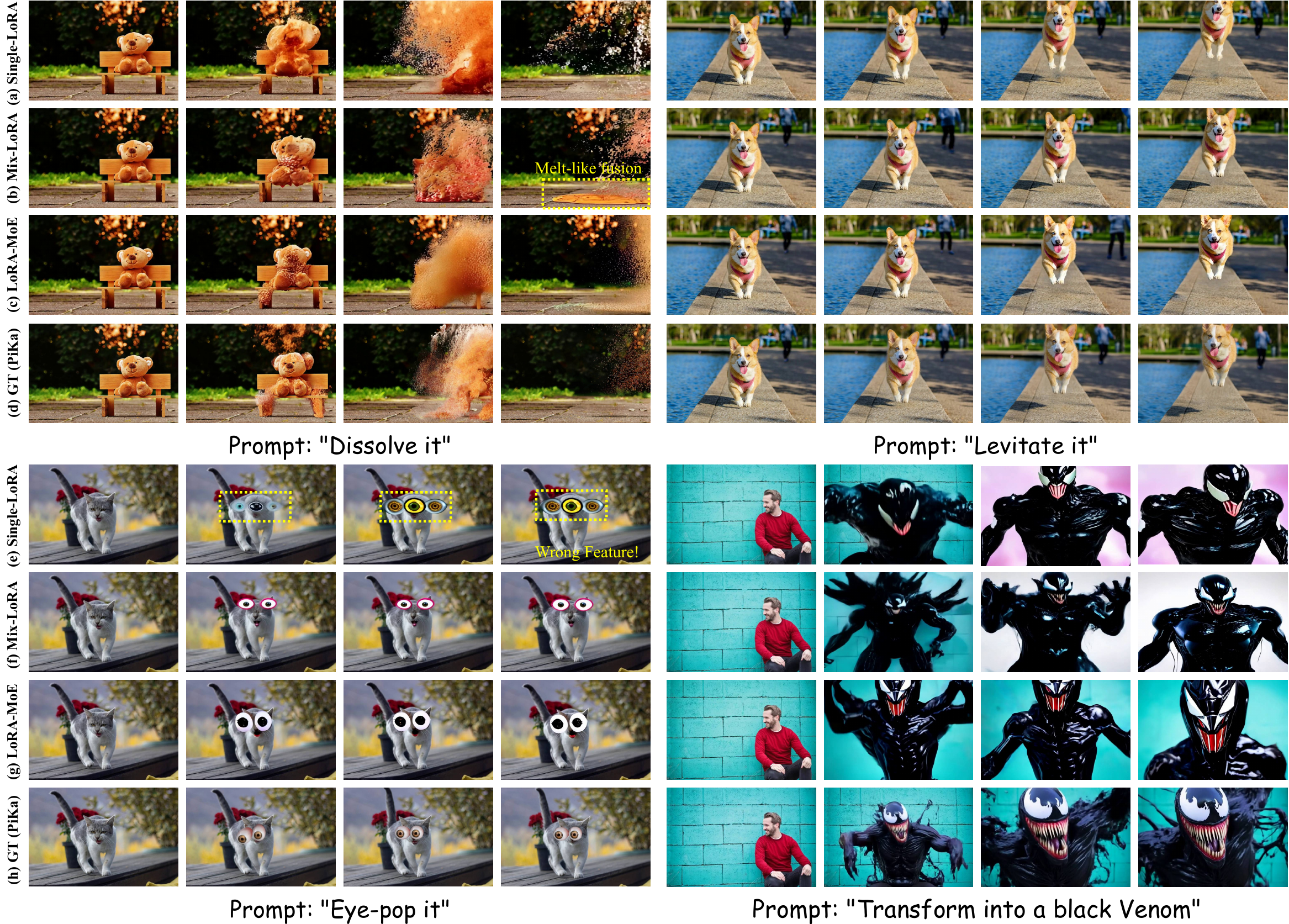}
    \caption{\textbf{Qualitative Comparison of Different LoRA Settings.}}
    \label{fig:qualitative-LoRAMoE}
\end{figure*}

\section{Metrics}

Our evaluation framework comprises three components: VFX detection, VFX controllable estimation, and video motion estimation, with the specific architecture illustrated in the Figure~\ref{fig:metrics}. These components correspond to the calculation processes of the metrics \emph{Effect Occurrence Rate (EOR)}, \emph{Effect Controllability Rate (ECR)}, and \emph{Regional Dynamic Degree (RDD)}.

\subsection{Visual Effects Detection}
To determine whether the desired visual effect is present in the generated videos, we leverage the Gemini 2.5~\cite{comanici2025gemini} large multimodal model as an evaluation assistant. Specifically, for each test video, we pair it with a predefined prompt template that explicitly describes the target effect, and input both into Gemini. We then request Gemini to provide a binary answer as to whether the specified effect has appeared in the video. Each query is repeated three times, and the most frequently occurring answer among the three is selected as the final result.

By applying this process to all videos in the evaluation set, we compute the \emph{Effect Occurrence Rate (EOR)}, defined as the percentage of videos in which Gemini confirms the occurrence of the designated visual effect.

\subsection{Visual Effects Controllability Estimation}
Controllability estimation is performed only on videos that are identified as having the target effect by the Visual Effects Detection step. For these videos, we compute the absolute pixel-wise difference between the first and last frames and compute the resulting difference binary map. In the masked region, we select the top 80\% of difference values; for the non-masked region, the bottom 80\%. The mean squared error (MSE) is then computed as \textit{Inner Diff} and \textit{Outer Diff}. An effect is regarded as controllable if \textit{Inner Diff} is below a threshold of 0.5 and \textit{Outer Diff} is below a threshold of 0.1. The \emph{Effect Controllability Rate (ECR)} is defined as the fraction of detected-effect videos that meet these criteria.

\subsection{Video Motion Detection}
The \emph{Regional Dynamic Degree (RDD)} measures the strength of motion caused by VFX within mask-specified region of a video. We use the RAFT algorithm~\cite{teed2020raft} to estimate optical flow between consecutive frames. Given a binary mask $\boldsymbol{s}$ for the region of interest, we calculate the mean motion magnitude within this region as follows:
\begin{equation}
\mathrm{RDD} = \frac{1}{N} \sum_{t=1}^{N} \frac{1}{|\boldsymbol{s}|} \sum_{(x, y) \in \boldsymbol{s}} \left\| \mathrm{Flow}(I_t, I_{t+1})_{x,y} \right\|_2,
\end{equation}
where $N$ is the number of consecutive frame pairs and $|\boldsymbol{s}|$ is the number of pixels in the masked region. A higher RDD indicates stronger or more dynamic effects within the specified area, enabling precise, region-specific evaluation of effect intensity.

\begin{table*}[!t]
\centering
\begin{tabular}{ccccccccccc}
\toprule
 & \textbf{Method} & \begin{tabular}[c]{@{}c@{}}\textbf{VFX}\\ \textbf{Nums}\end{tabular} & \textbf{Deflate} & \textbf{Melt} & \textbf{Crumble} & \textbf{Dissolve} & \textbf{Ta-da} & \textbf{Squish} & \textbf{Crush} & \textbf{Cake-ify} \\ \midrule
\multirow{6}{*}{\textbf{FVD}$\downarrow$} & \multirow{6}{*}{\textbf{LoRA}} & 1 & 913 & \underline{2827} & 2947 & 1770 & \textbf{1053} & \textbf{1415} & \textbf{1496} & 2138 \\
 &  & 2 & 857 & 3606 & - & - & - & - & - & - \\
 &  & 4 & \textbf{779} & 3957& \textbf{2324} & \textbf{1559} & - & - & - & - \\
 &  & 8 & \underline{787} & 3456 & 2756 & 2103 & \underline{1206} & 1699 & \underline{1540} & \textbf{1355} \\
 &  & 14 & 949 & \textbf{2030} & \underline{2718} & \underline{1722} & 1501 & \underline{1527} & 1570 & \underline{1388} \\ \midrule
 & \textbf{LoRA MoE} & 14 & 839 & 2512 & \textbf{1641} & \textbf{1118} & 1064 & 1561 & \textbf{1213} & 1506 \\
 \bottomrule
\end{tabular}
\caption{\textbf{Ablation Study on Co-trained with Different VFX Number.}}
\label{tab:co-train}
\vspace{-10pt}
\end{table*}

\section{Experiments Results Details}

\subsection{Qualitative Results of LoRA MoE}

Qualitative results of Different LoRA settings are visualized in Figure~\ref{fig:qualitative-LoRAMoE}. LoRA-MoE demonstrates superior visual performance.

\subsection{Qualitative Results of single-VFX control}

\begin{figure}[!h]
    \centering
    \includegraphics[width=1\linewidth]{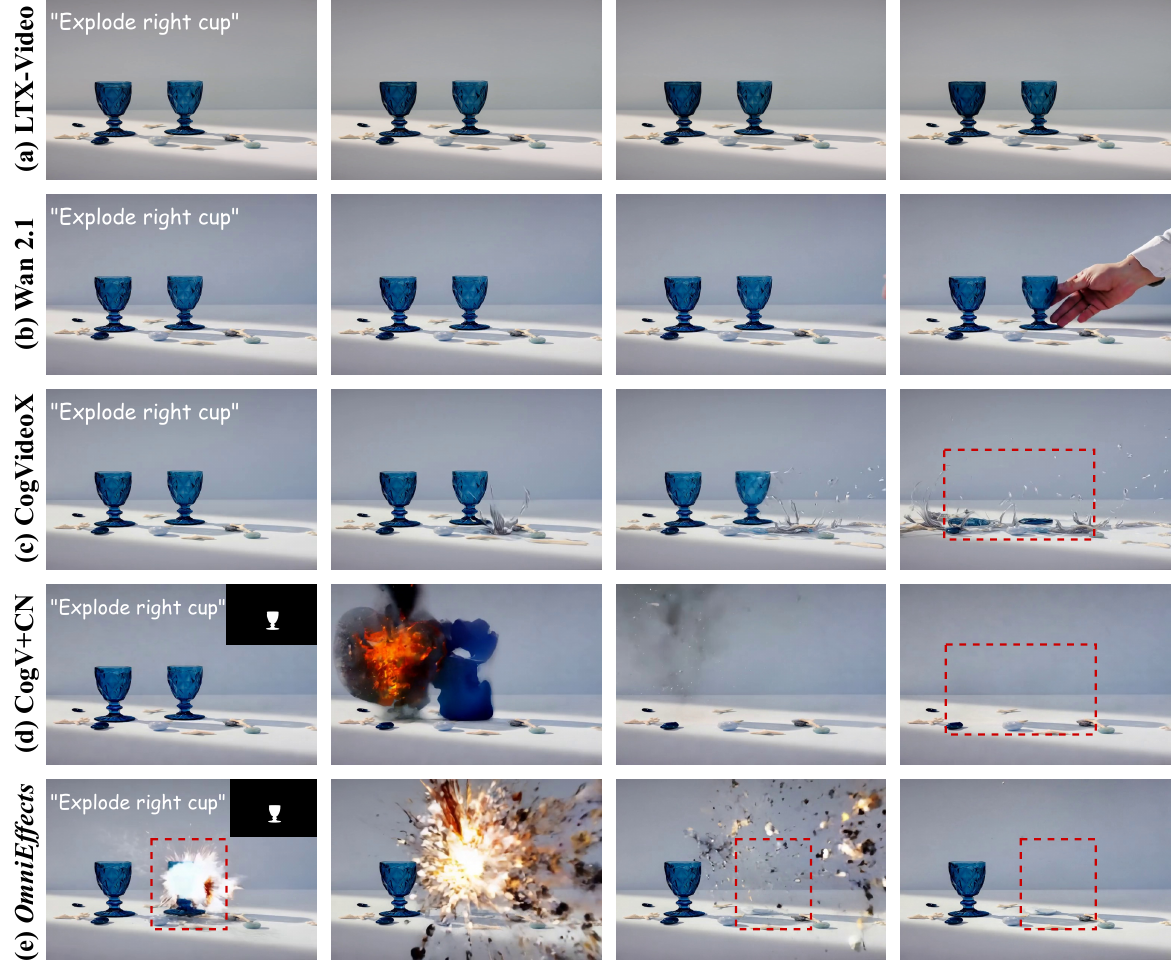}
    \caption{\textbf{Qualitative Comparison of Single-VFX Generation.} The desired outcome requires the right cup to explode
while the left stays static.}
    \label{fig:single-vfx-qualitative}
\end{figure}

Qualitative results of single-VFX control are visualized in Figure~\ref{fig:single-vfx-qualitative}. CogV+CN incorrectly causes both cups to explode, while our proposed \textit{Omni-Effects} explodes the correct cup while keeping the other cup intact.

\subsection{User Study of Multi-VFX Generation }
\begin{figure}[!ht]
    \centering
     \includegraphics[width=1\linewidth]{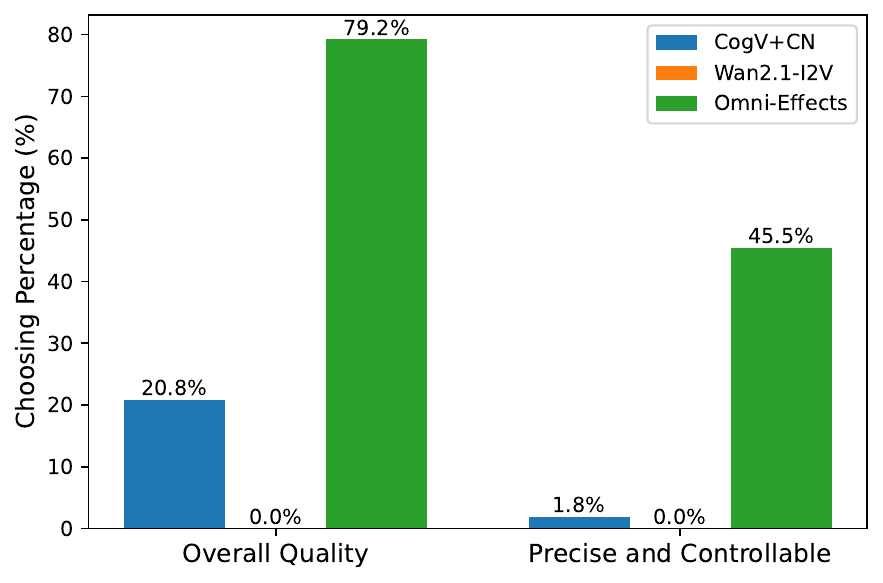}
    \caption{\textbf{User Study for Multi-VFX Generation.} \textit{Omni-Effects} exceeds other baseline.}
    \label{fig:userstudy}
\end{figure}

To achieve more reliable evaluation results, we select a subset of videos from the test set for a user study. We choose the state-of-the-art open-source Wan2.1-I2V model as the representative base model, and evaluate it alongside \textbf{CogV+CN} and our proposed method on multi-effect videos. We design two questions: one regarding user preference (i.e., \textit{which video the user considers to be of the highest overall quality?}), and another asking whether each video demonstrates precise controllability of the specified visual effect, which can directly reflect the effectiveness of our approach. Six professional raters participated in the evaluation, and the results are shown in Figure~\ref{fig:userstudy}

\section{More Ablation Study}

\subsection{VFX-combination training}

Ablation studies on diverse VFX-combination training regimes (Table~\ref{tab:co-train} and Table~\ref{tab:deflate}) reveal inherent effect clustering, demonstrating that compatible VFX-combination training enhances single-VFX generation quality. The proposed LoRA-MoE framework effectively leverages this property to boost performance across all VFX types.

\begin{table}[!t]
\centering
\begin{tabular}{cccc}
\toprule
 & \textbf{Deflate} & \textbf{Deflate+Melt} & \textbf{Deflate+Squish} \\ \midrule 
\textbf{FVD}$\downarrow$ & 913 & 857($\downarrow6\%$) & 1611($\uparrow76\%$) \\ 
\bottomrule
\end{tabular}
\caption{\textbf{Ablation Study on Co-trained with Different VFX combination.}}
\label{tab:deflate}
\end{table}

\subsection{Data Augmentation}

Ablation study on data augmentation (Figure~\ref{fig:schedule}) reveals that models struggle to achieve spatial controllability using single-VFX data alone without augmentation. Our custom-designed data augmentation protocol enables effective controllability for both single- and multi-VFX generation.

\subsection{Timestep Scheduler}

\begin{figure}[!t]
    \centering
    \includegraphics[width=1\linewidth]{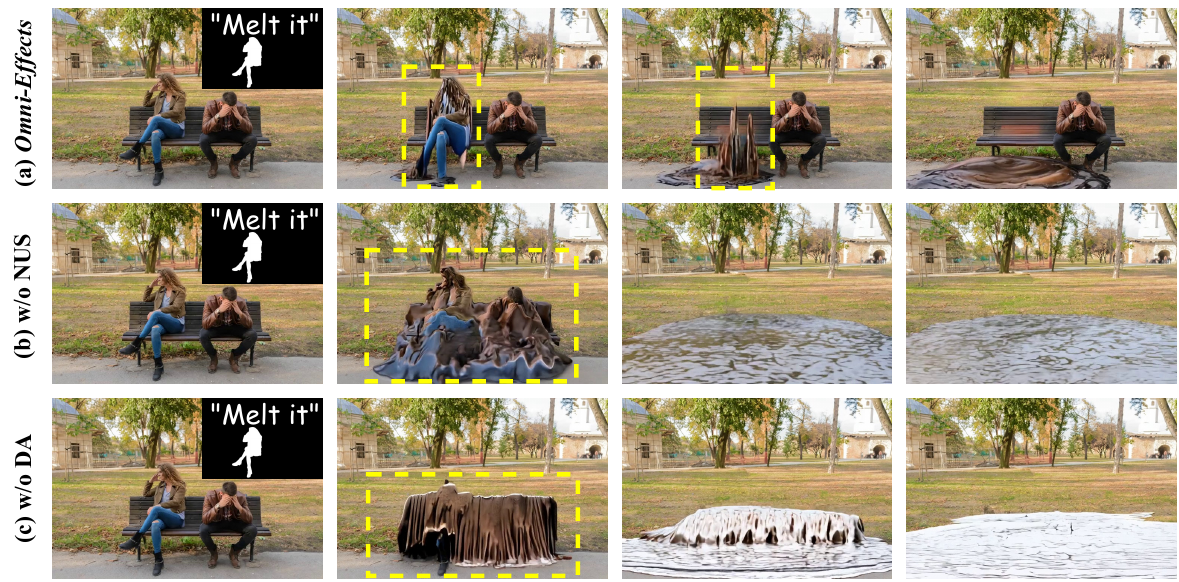}
    \caption{\textbf{Ablation study on Non-uniform Timestep Sampling Steps.} \textbf{NUS} stands for Non-Uniform Sampling, and \textbf{DA} stands for Data Augmentation. The training of first stage $N=1$ with epoch=70.}
    \label{fig:schedule}
\end{figure}

Ablation study on different time schedulers (Figure~\ref{fig:schedule}) reveals that traditional uniform sampling necessitates extensive training efforts to sufficiently optimize model accuracy in initial steps, whereas our employed non-uniform timestep sampling empirically accelerates model convergence.

\subsection{Training Strategy}

\begin{figure}[!t]
    \centering
    \includegraphics[width=1\linewidth]{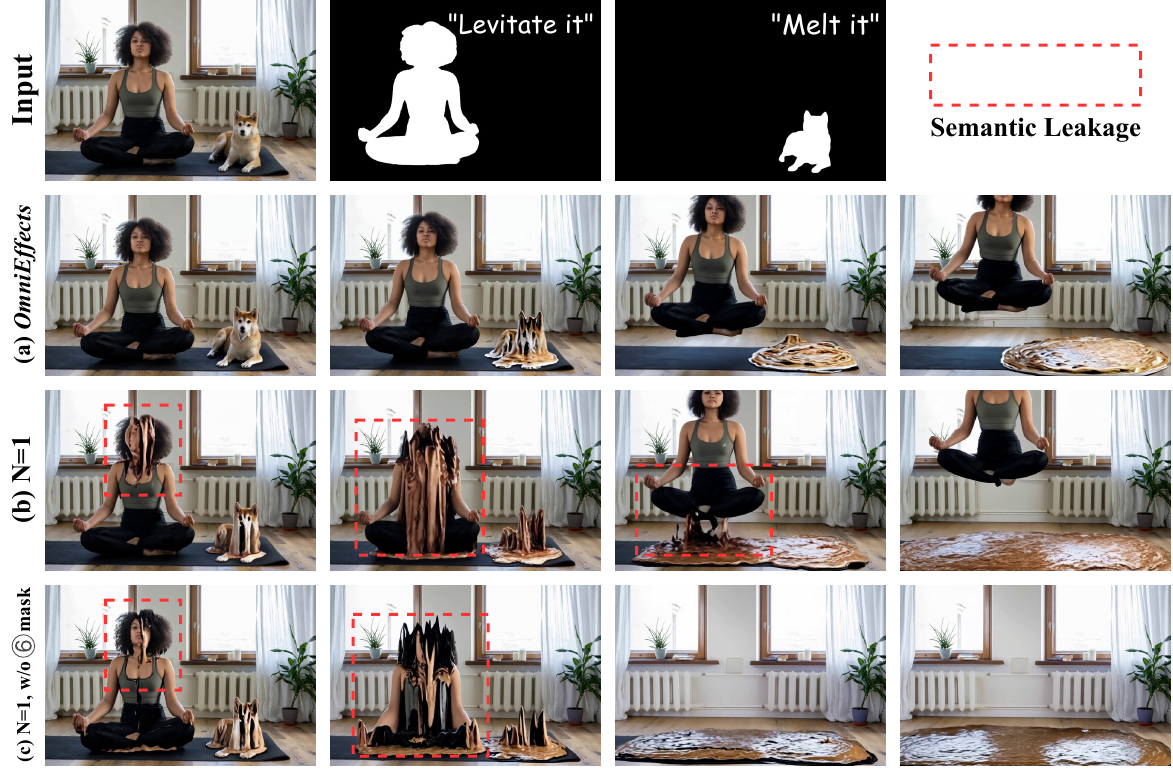}
    \caption{\textbf{Ablation Study on Different Training Strategy.}}
    \label{fig:training-n}
\end{figure}

Ablation study on different training strategies (Figure~\ref{fig:training-n}) reveals that our employed Dual-phase Training Strategy effectively enhances model robustness for multi-VFX generation, with Stage 2 demonstrably mitigating interference artifacts compared to Stage 1.

\section{Limitation}

As the number of combined VFX categories $N$ increases (especially when $N>2$), the performance tends to decrease. We attribute this phenomenon to the lack of real multi-VFX data, which leads to a bias between training and inference. In future work, we plan to collect more high-quality multi-VFX data to improve the stability of our method.

\end{document}